\documentclass[conference]{ieeeconf}
\IEEEoverridecommandlockouts                          
\usepackage{amsmath,amssymb,amsfonts}
\usepackage{graphicx}
\usepackage{textcomp}
\usepackage{xcolor}
\usepackage{booktabs}
\usepackage{multirow}
\usepackage{algorithm}
\usepackage{algpseudocode}
\usepackage{cite}
\let\labelindent\relax  
\usepackage{enumitem}
\usepackage{placeins}
\usepackage{afterpage}  
\usepackage{siunitx}
\usepackage[colorlinks=true,urlcolor=blue,linkcolor=black,citecolor=black,bookmarks=true]{hyperref}

\title{\LARGE \bf Learnable Conformal Prediction with Context-Aware Nonconformity Functions for Robotic Planning and Perception}

\author{Divake Kumar$^1$, Sina Tayebati$^1$, Francesco Migliarba$^1$, Ranganath Krishnan$^2$, Amit Ranjan Trivedi$^1$ \\
$^1$University of Illinois at Chicago, $^2$Intel Labs}

\begin{document}

\maketitle
\thispagestyle{empty}
\pagestyle{empty}

\begin{abstract}
Deep learning models in robotics often output point estimates with poorly calibrated confidences, offering no native mechanism to quantify predictive reliability under novel, noisy, or out-of-distribution inputs. Conformal prediction (CP) addresses this gap by providing distribution-free coverage guarantees, yet its reliance on fixed nonconformity scores ignores context and can yield intervals that are overly conservative or unsafe. We address this with Learnable Conformal Prediction (LCP), which replaces fixed scores with a lightweight neural function $s_\theta(x) = f_\theta(\phi(x))$ that leverages geometric, semantic, and model cues. Trained to balance coverage, efficiency, and calibration, LCP preserves CP’s finite-sample guarantees while producing intervals that adapt to instance difficulty, achieving context-aware uncertainty without ensembles or repeated inference. On the MRPB benchmark, LCP raises navigation success to 91.5\% versus 87.8\% for Standard CP, while limiting path inflation to 4.5\% compared to 12.2\%. For object detection on COCO, BDD100K, and Cityscapes, it reduces mean interval width by 46--54\% at 90\% coverage, and on classification tasks (CIFAR-100, HAM10000, ImageNet) it shrinks prediction sets by 4.7--9.9\%. The method is also computationally efficient, achieving real-time performance on resource-constrained edge hardware (Intel NUC with footprint $4.6 \times 4.4$ inch$^2$ and power $<30$\,W) while simultaneously providing uncertainty estimates along with the mean prediction.

The code and demonstration videos are available at: \href{https://divake.github.io/learnable-cp-robotics/}{\textit{https://divake.github.io/learnable-cp-robotics/}}
\end{abstract}

\section{Introduction and Prior Works}
Learning from data is an inherently ill-conditioned problem that often admits multiple optimal solutions. Selecting a single solution while discarding others is theoretically unjustified and thus limits predictive robustness. Consequently, most learning models that output point predictions or poorly calibrated confidences \cite{guo2017calibration} incur prediction errors that depend heavily on context such as occlusion, clutter, or distribution shift \cite{taori2020shift}. Moreover, these models are typically optimized for average-case accuracy rather than worst-case reliability in deployment \cite{amodei2016concrete}. As a result, while they may deliver strong mean performance, yet can fail catastrophically in rare yet safety-critical corner cases, resulting in a crucial limitation for their deployment for mission/safety-critical robotics.

Two primary sources of uncertainty exist in learning models: \textit{epistemic uncertainty}, arising from limited data or model capacity and reducible with additional information, and \textit{aleatoric uncertainty}, caused by inherent sensor noise or environmental ambiguity which is irreducible even with infinite data \cite{der2009aleatoric, kendall2017uncertainties}. Recent advances have shown promise in autonomous navigation~\cite{zhang2024conformal,luo2024efficient} and human-robot collaboration~\cite{park2024adaptive}. Recent work has explored separating these uncertainties through conformal inference and evidential learning \cite{stutts2024conformal}, enabling more nuanced risk assessment. A range of uncertainty quantification (UQ) methods aim to capture these effects. Bayesian neural networks and variational inference estimate epistemic uncertainty but require multiple stochastic passes, making them impractical for time-constrained control \cite{blundell2015weight}. Deep ensembles provide stronger calibration but are computationally expensive \cite{lakshminarayanan2017simple}. Approximate approaches such as Monte Carlo dropout reduce overhead but often yield poorly calibrated estimates \cite{gal2016dropout}, while post-hoc calibration adjusts confidence scores without statistical guarantees \cite{guo2017calibration}. 

\begin{figure}[t]
\centering
\includegraphics[width=\linewidth]{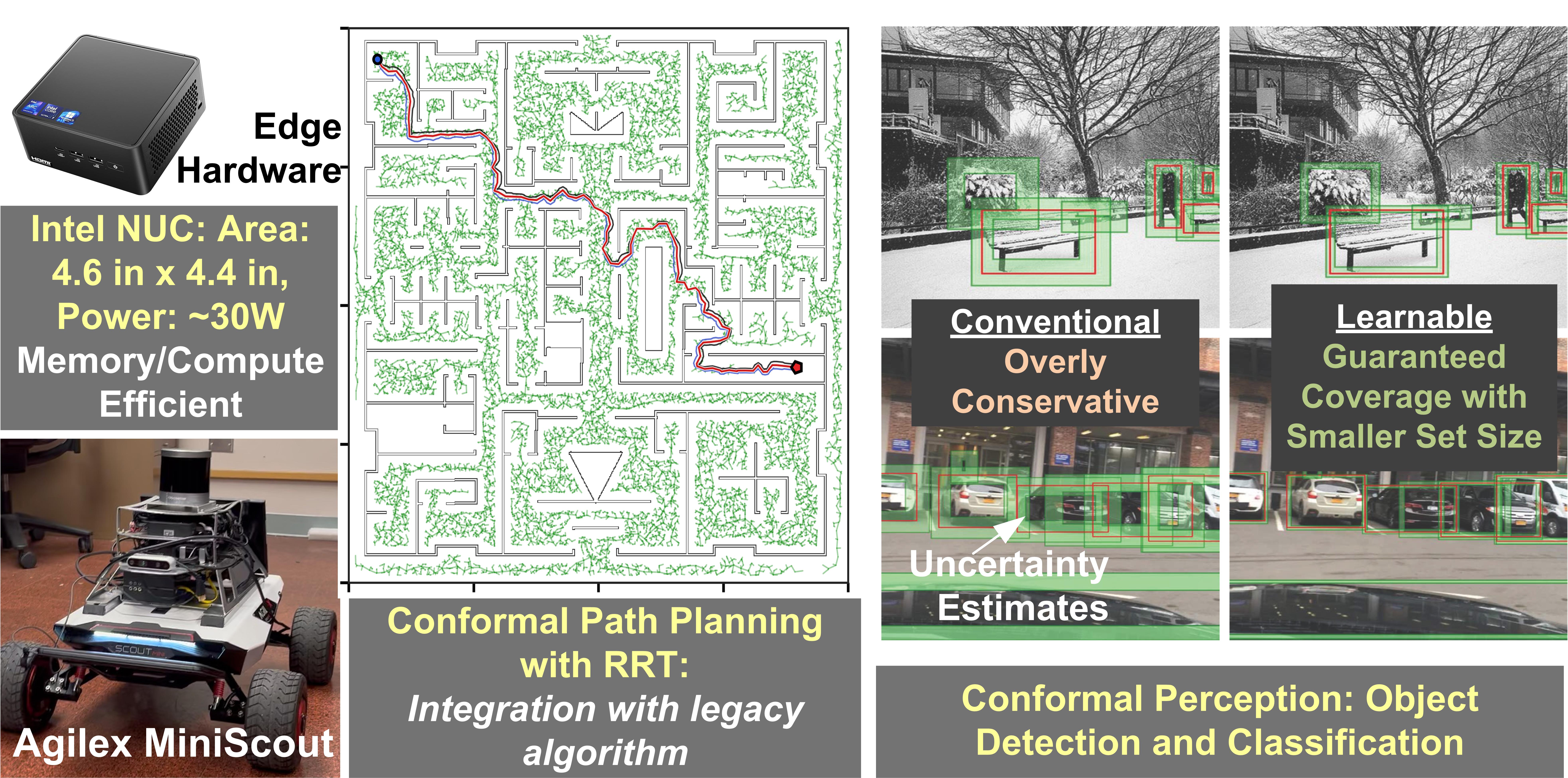}
\vspace{-20pt}\caption{\textbf{Real-time conformal prediction for safe and efficient robotics:} 
\textbf{Left:} The framework runs efficiently on an Intel NUC (4.6\,in $\times$ 4.4\,in, $<30$\,W) mounted on an Agilex MiniScout platform. 
\textbf{Center:} Conformal path planning integrates with legacy algorithms such as RRT. 
\textbf{Right:} Conformal perception improves object detection and classification, achieving guaranteed coverage with smaller set sizes.}
\label{fig:vision}
\end{figure}

Conformal prediction (CP) has recently attracted significant interest as a principled framework for uncertainty quantification \cite{vovk2005, angelopoulos2023conformal, stutts2025uncertainty,stutts2024mutual, stutts2023lightweight}, with recent extensions to multi-sensor fusion \cite{kumar2025uncertainty} and adaptive abstention policies \cite{tayebati2025learningconformal}. Originating in statistical learning theory, CP constructs prediction sets calibrated on held-out data and provides distribution-free, finite-sample coverage guarantees under the assumption of exchangeability. Unlike many heuristic post-hoc calibration methods, CP offers explicit statistical guarantees, ensuring that true outcomes fall within the predicted sets at a user-specified confidence level. Moreover, CP is suited even for legacy prediction models that do not necessarily rely on learning from data. These properties make CP particularly appealing for robotics, where models must operate under distribution shift and safety requires formal reliability bounds on decision-making.

Despite its generality, CP is most often implemented with fixed nonconformity functions that fail to capture the intense interaction of input data, application domain, and context in shaping uncertainty. For instance, in regression, standard CP with residual-based nonconformity produces intervals of constant width across all inputs, ignoring heteroscedasticity \cite{papadopoulos2002inductive,lei2018distribution}. As a result, CP provides valid coverage but does not account for how uncertainty emerges from the joint structure of observations and operating conditions. This limitation is especially critical in robotics, where risk depends not only on the raw input but also on situational context: for example, a partially occluded object may be harmless clutter in a warehouse aisle yet represent a pedestrian entering a crosswalk in an urban scene. Treating both as equally uncertain either wastes efficiency in benign settings or under-protects in safety-critical ones.

We address this limitation with \textit{Learnable Conformal Prediction (LCP)} (Fig. \ref{fig:vision}). Instead of fixed scores, we introduce a feature-driven function $s_\theta(x) = f_\theta(\phi(x))$ that adapts to the structure of prediction errors. Features $\phi(x)$ encode geometric, semantic, and model-derived cues, while $f_\theta$ is a lightweight neural network trained to balance coverage, efficiency, and calibration. Calibration over these learned scores preserves the finite-sample coverage guarantees of CP \cite{vovk2005,lei2018distribution}, while producing intervals that shrink in simple cases and expand in difficult ones. 


We evaluate LCP for (i) robotic path planning under noisy and incomplete sensing on the MRPB benchmark, (ii) object detection with uncertainty calibration on COCO, BDD100K, and Cityscapes, and (iii) image classification on CIFAR-100, HAM10000, and ImageNet. Across these benchmarks, LCP consistently improves the safety--efficiency trade-off across planning, perception, and classification tasks. On the MRPB path-planning benchmark, LCP raises success rates to 91.5\% while limiting path inflation to 4.5\%, compared to 87.8\% success and 12.2\% inflation with standard CP. For object detection on COCO, BDD100K, and Cityscapes, LCP reduces mean interval width by 46--54\% while sustaining $\approx$90\% coverage. In classification (CIFAR-100, HAM10000, ImageNet), it cuts prediction set sizes by 4.7--9.9\% relative to fixed baselines without losing validity. The proposed framework is also computationally efficient, achieving real-time performance on resource-constrained edge hardware (Intel NUC, area $4.6 \times 4.4$\,inch$^2$, power $<30$\,W) while simultaneously extracting uncertainty estimates and prediction. This aligns with recent advances in edge robotics \cite{darabi2024navigating} and intelligent sensing-to-action systems \cite{trivedi2025intelligent}.

\section{Learnable Conformal Prediction (LCP) by training Nonconformity Scoring Function}
\label{sec:method}
Conformal prediction (CP) is a distribution-free framework for constructing statistically valid prediction sets. Given a calibration dataset $\{(x_i, y_i)\}_{i=1}^n$ and a new test input $x_{n+1}$, the goal is to form a set $C(x_{n+1})$ that contains the true label $y_{n+1}$ with probability at least $1-\alpha$. Formally, under exchangeability, CP guarantees
\(
\mathbb{P}(Y \in C(X)) \geq 1-\alpha
\)
\cite{vovk2005algorithmic}.  

This guarantee is obtained by calibrating a nonconformity score $s(x,y)$, which measures how unusual a candidate label $y$ is for input $x$. The prediction set contains labels with scores below the empirical quantile $\hat{q}$ from the calibration set:
\[
C(x) = \{ y \in \mathcal{Y} \mid s(x,y) \leq \hat{q} \}, \]
\[
\hat{q} = \text{Quantile}\!\left(\{s(x_i, y_i)\}_{i=1}^n, \tfrac{\lceil(n+1)(1-\alpha)\rceil}{n}\right).
\]

While validity holds regardless of $s$, the usefulness of CP depends on the trade-off between prediction set size and informativeness. The expected size $\mathbb{E}[|C(X)|]$ is determined by both the base model’s accuracy and the choice of $s(x,y)$. In practice, set size governs the safety–efficiency balance: larger sets provide more caution but reduce decisiveness, while smaller sets are efficient but risk undercoverage. Thus, the scoring function is a \textit{key lever} for navigating this trade-off without compromising statistical rigor.

\subsection{Limitations of Classical Nonconformity Scoring}
Most CP methods use simple, hand-crafted rules that convert model outputs into nonconformity scores, such as:  

\begin{itemize}[leftmargin=*, itemsep=1pt, topsep=2pt]
    \item \textbf{Probability-based:} Scores such as $s(x,y) = 1 - p(y|x)$ assign low values to high-probability classes. Adaptive Prediction Sets (APS)~\cite{romano2019conformalized} instead accumulate the probability mass of higher-ranked labels.  
    \item \textbf{Margin-based:} Gap-based rules like $s(x,y) = \log(\max_{j} p(j|x)) - \log(p(y|x))$ measure separation between the top class and candidate $y$.  
    \item \textbf{Logit-based:} Methods such as Sparsemax~\cite{martins2016softmax} operate in logit space by projecting onto a sparse simplex.  
\end{itemize}

These scores are computationally efficient but rigid. For instance, softmax captures confidence yet ignores factors such as occlusion or distribution shift, often producing sets that are over-conservative in easy and unsafe in difficult ones. Recent work on real-time motion planning~\cite{fan2024realtime} and probabilistic safety~\cite{sun2024probabilistic} highlights the need for context-aware uncertainty quantification.

\subsection{Learning Nonconformity Scoring from Data}

We introduce \emph{learnable} nonconformity functions that adapt to the base model and dataset. Instead of prescribing a fixed formula, we parameterize $s_\theta(x,y)$ and train it to yield scores that are valid, efficient, and well-calibrated. We outline the approach for three robotics use-cases:

\noindent\textbf{Path planning.}
For each waypoint $w$, we construct a 20-dimensional feature vector $\phi(w)$ capturing geometry, uncertainty, and local context. Geometric features include minimum clearance $d_{\text{min}}(w,\hat{\mathcal{O}})$, average clearance at radii $\{1,2\}$\,m, passage width (largest inscribed circle), and obstacle density.  Path context features include normalized progress $i/|p|$, distance to goal, curvature $\kappa(w)=\|\ddot{p}(s)\|$, velocity, and heading change $\Delta\theta$. An MLP (128-64-32-1, BatchNorm, 20\% dropout) maps $\phi(w)$ to an adaptive margin:
\begin{equation}
\tau(w)=f_\theta(\phi(w)), \qquad 
\tau_{\text{final}}(w)=\max(r,\, \tau(w)+q^\ast),
\label{eq:path_tau}
\end{equation}
$r=0.17$\,m is robot radius and $q^\ast$ is the calibrated offset.

Training of the nonconformity function $s_\theta(x)$ balances coverage, efficiency, and task-specific goals through tailored objectives. An asymmetric Huber loss penalizes unsafe margins more heavily:
\begin{equation}
\mathcal{L}_{\text{safety}} =
\begin{cases}
0.5 \cdot \text{Huber}(\tau - d, 0), & \tau \ge d, \\
2.0 \cdot \text{Huber}(\tau - d, 0), & \tau < d,
\end{cases}
\label{eq:safety_loss}
\end{equation}
and the full path-planning loss adds efficiency, smoothness, and coverage terms:
\begin{equation}
\mathcal{L}_{\text{path}} = \mathcal{L}_{\text{safety}}
+ 0.3\|\tau - 0.3\|
+ 0.2\sum_i (\tau_{i+1}-\tau_i)^2
+ \mathcal{L}_{\text{coverage}},
\label{eq:path_loss}
\end{equation}
where the smoothness term discourages abrupt changes.

\noindent\textbf{Object detection.}
We also evaluated learnable nonconformity functions for object detection. Here, each bounding box $b$ is represented by a 13-dimensional vector $\phi(b)$ including normalized coordinates, detector confidence, log area, aspect ratio, and distance to the image center. Coverage targets are scaled by size:
\begin{equation}
\text{target}_{\text{cov}} =
\begin{cases}
0.90, & \sqrt{\text{area}}<32, \\
0.89, & 32\le \sqrt{\text{area}}<96, \\
0.85, & \sqrt{\text{area}}\ge 96,
\end{cases}
\label{eq:size_targets}
\end{equation}
so smaller objects receive higher coverage. For the nonconformity scoring $s_\theta(x)$, a network (256-128-64-4, ELU) outputs symmetric interval widths 
$\mathbf{w}=[w_{x_0},w_{y_0},w_{x_1},w_{y_1}]$, later scaled by a calibrated factor $\tau$.

To train $s_\theta(x)$, we minimize the Mean Prediction Interval Width (MPIW), normalized by box size:
\begin{equation}
\mathrm{MPIW} = \tfrac{1}{4}\sum_{j \in \{x_0,y_0,x_1,y_1\}} 2w_j\tau, 
\mathcal{L}_{\text{MPIW}} = \frac{\mathrm{MPIW}}{(w_{\text{box}} + h_{\text{box}})/2}.
\label{eq:mpiw}
\end{equation}
A coverage penalty enforces the target range $[0.88,0.92]$:
\begin{equation}
\mathcal{L}_{\text{penalty}} =
\begin{cases}
5(\hat{C}-0.89)^2, & \hat{C}>0.905, \\
10(0.89-\hat{C})^2, & \hat{C}<0.88, \\
0, & \text{otherwise}.
\end{cases}
\label{eq:coverage_penalty}
\end{equation}

\noindent\textbf{Classification.}
Extending the evaluations of $s_\theta(x)$ for classification, we extract the following features for each class $c$: probability $p(c|x)$, normalized rank $\text{rank}(p_c)/K$, margin to the top class, top-$k$ indicators for $k\in\{1,3,5\}$, entropy contribution $-p(c|x)\log p(c|x)$, and global maximum probability. The multi-layer perception width for $s_\theta(x)$ scales with the number of classes to balance capacity and regularization:
\begin{equation}
(H_1,H_2)=
\begin{cases}
(32,16), & K\le 10, \\
(64,32), & 10<K\le 100, \\
(128,64), & 100<K\le 1000, \\
(256,128), & K>1000.
\end{cases}
\label{eq:arch_scaling}
\end{equation}

A three-phase schedule guides training: 

\noindent\underline{Phase 1} (epochs 1–10) uses margin loss:
\begin{equation}
\mathcal{L}_{\text{margin}} = \text{ReLU}(s_{\text{true}} - \text{mean}(s_{\text{false}}) + \delta), \quad \delta=0.8.
\label{eq:margin_loss}
\end{equation}

\noindent\underline{Phase 2} (epochs 11–20) adds coverage 
$\mathcal{L}_{\text{cov}} = (\hat{C}-(1-\alpha))^2$,  
\noindent\underline{Phase 3} (epochs 21+) introduces set size minimization:
\begin{equation}
\mathcal{L}_{\text{size}} = \frac{1}{n}\sum_{i=1}^n \frac{|C(x_i)|}{K}
+ \lambda_{\text{empty}}\mathbf{1}[|C(x_i)|=0].
\label{eq:size_loss}
\end{equation}

\subsection{Calibration and Adaptive Thresholds}

Post-training calibration restores coverage guarantees while maintaining efficiency. For path planning, we compute an additive offset:
\begin{equation}
q^\ast = \text{Quantile}_{1-\alpha}\left(\{\tau_{\text{pred}}(w_i) - d_{\text{true}}(w_i)\}_{i=1}^m\right).
\label{eq:path_calib}
\end{equation}
For object detection, calibration uses a multiplicative factor based on the infinity norm of prediction errors:
\begin{equation}
\tau = \text{Quantile}_{1-\alpha}\left(\left\{\frac{\|\mathbf{b}^\ast_i - \hat{\mathbf{b}}_i\|_\infty}{f_\theta(\phi(\hat{\mathbf{b}}_i))}\right\}_{i=1}^m\right).
\label{eq:detect_calib}
\end{equation}
During training, thresholds are updated with an exponential moving average:
\begin{equation}
\tau_t = \beta\tau_{t-1} + (1-\beta)\hat{q}_t, \quad \beta = 0.95,
\label{eq:tau_ema}
\end{equation}
which balances stability ($\beta$ close to 1) with adaptability to distribution shifts. For classification, we apply smoothed quantile estimation with asymmetric windows:
\begin{equation}
\hat{q} = \frac{\sum_{i=k-3}^{k+1} w_i s_i}{\sum_{i=k-3}^{k+1} w_i}, \quad w_i = 1.5 - 0.1|i-k|,
\label{eq:smooth_quantile}
\end{equation}
where $k$ indexes the $(1-\alpha)$ quantile. The asymmetric window, with greater weight below the quantile, yields conservative yet stable estimates.

\subsection{Optimization and Implementation Details}
We optimize training of nonconformity scoring functions ($s_\theta(x)$) using AdamW with weight decay $\lambda_{\text{wd}}=10^{-5}$ and a cosine annealing schedule. For classification, we apply CosineAnnealingWarmRestarts with $T_0=5$ epochs and $\eta_{\min}=10^{-5}$. Gradient clipping at norm 0.5 prevents early-stage instability when scores are poorly calibrated. Batch sizes are task-specific: 256 for classification (stable statistics), 512 for detection (memory–efficiency trade-off), and 1024 for path planning (trajectory-level parallelism).

Loss weights adapt dynamically to coverage performance. When empirical coverage $\hat{C} < 1-\alpha-\epsilon$ with $\epsilon=0.02$, coverage weight increases ($w_c=2.0$, $w_s=1.0$). Otherwise, efficiency is prioritized ($w_c=1.0$, $w_s=1.5$). This guarantees coverage before minimizing set sizes. For detection, size-stratified metrics are also monitored, and weights are adjusted per category to balance performance across scales.

Running statistics are maintained per feature dimension, and inputs are standardized as $\tilde{\phi}_i=(\phi_i-\mu_i)/\sigma_i$. Path planning uses per-environment normalization to account for map variation, object detection uses global dataset statistics, and classification applies per-dataset normalization to handle differing class distributions. 


\subsection{Computational Complexity and Efficiency}

Despite a more elaborate scoring function design, the overhead of learnable CP remain minimal relative to the base model. Feature extraction costs $O(d)$ per instance, where $d$ (8–20 depending on task) is the feature dimension. The MLP forward pass requires $O(H_1d + H_1H_2 + H_2)$ operations, with $H_1,H_2$ denoting hidden layer sizes. Even for the largest network (256–128 neurons), this adds under 0.5\,ms on GPU and under 5\,ms on CPU per prediction.

Memory use is also modest. The largest model (classification on ImageNet) stores only 100\,KB of parameters, while path planning models require about 42\,KB. Feature caching during training costs $O(nd)$, where $n$ is the dataset size, but inference needs only $O(1)$ memory per instance. This efficiency is crucial for deployment on emerging probabilistic hardware architectures \cite{lee2024highly}. Calibration involves sorting $n$ scores ($O(n\log n)$) once during setup, with no additional overhead at deployment.

\begin{table*}[!t]
\setlength{\tabcolsep}{3.5pt}
\centering
\textbf{Table I: Path Planning Results on MRPB across Five Environments Comparing Naive, Standard CP, and LCP.}\vspace{2pt}
\label{tab:path_planning_results}
\begin{tabular}{llcccccccc}
\toprule
\textbf{Environment} & \textbf{Method} & \textbf{Success Rate ↑} & \textbf{Path Length (m) ↓} & \textbf{Waypoints ↓} & \textbf{$d_0$ (m) ↑} & \textbf{$d_{avg}$ (m) ↑} & \textbf{$p_0$ (\%) ↓} & \textbf{T (s) ↓} \\
\midrule
\multirow{3}{*}{\textbf{office01add}} 
& Naive & 81.62\% & \textbf{21.92 $\pm$ 2.15} & \textbf{35 $\pm$ 3} & 0.212 $\pm$ 0.048 & 0.637 $\pm$ 0.085 & 4.92 $\pm$ 1.23 & \textbf{18.54 $\pm$ 2.35} \\
& Standard CP & 89.24\% & 24.79 $\pm$ 2.38 & 39 $\pm$ 4 & \textbf{0.342 $\pm$ 0.051} & \textbf{0.817 $\pm$ 0.092} & \textbf{1.48 $\pm$ 0.52} & 20.98 $\pm$ 2.65 \\
& \textbf{Learnable CP} & \textbf{92.78\%} & 22.82 $\pm$ 2.21 & 36 $\pm$ 4 & 0.285 $\pm$ 0.045 & 0.711 $\pm$ 0.088 & 2.56 $\pm$ 0.78 & 19.31 $\pm$ 2.42 \\
\midrule
\multirow{3}{*}{\textbf{office02}} 
& Naive & 77.18\% & \textbf{48.35 $\pm$ 4.62} & \textbf{85 $\pm$ 8} & 0.206 $\pm$ 0.052 & 0.649 $\pm$ 0.098 & 5.28 $\pm$ 1.65 & \textbf{42.65 $\pm$ 4.85} \\
& Standard CP & 87.56\% & 51.28 $\pm$ 4.85 & 89 $\pm$ 9 & \textbf{0.358 $\pm$ 0.058} & \textbf{0.832 $\pm$ 0.103} & \textbf{1.85 $\pm$ 0.68} & 45.23 $\pm$ 5.12 \\
& \textbf{Learnable CP} & \textbf{91.23\%} & 49.15 $\pm$ 4.71 & 86 $\pm$ 8 & 0.294 $\pm$ 0.052 & 0.744 $\pm$ 0.095 & 2.93 $\pm$ 0.95 & 43.35 $\pm$ 4.95 \\
\midrule
\multirow{3}{*}{\textbf{shopping\_mall}} 
& Naive & 75.64\% & \textbf{62.45 $\pm$ 5.38} & \textbf{55 $\pm$ 5} & 0.185 $\pm$ 0.045 & 0.753 $\pm$ 0.112 & 6.15 $\pm$ 1.82 & \textbf{45.32 $\pm$ 5.25} \\
& Standard CP & 86.81\% & 68.21 $\pm$ 5.95 & 65 $\pm$ 6 & \textbf{0.365 $\pm$ 0.062} & \textbf{0.900 $\pm$ 0.118} & \textbf{1.23 $\pm$ 0.48} & 49.50 $\pm$ 5.68 \\
& \textbf{Learnable CP} & \textbf{90.37\%} & 64.87 $\pm$ 5.52 & 60 $\pm$ 6 & 0.308 $\pm$ 0.055 & 0.815 $\pm$ 0.108 & 2.45 $\pm$ 0.88 & 47.08 $\pm$ 5.42 \\
\midrule
\multirow{3}{*}{\textbf{room02}} 
& Naive & 83.25\% & \textbf{20.14 $\pm$ 1.95} & \textbf{30 $\pm$ 3} & 0.233 $\pm$ 0.055 & 0.741 $\pm$ 0.089 & 3.85 $\pm$ 1.15 & \textbf{16.26 $\pm$ 2.15} \\
& Standard CP & 90.43\% & 22.15 $\pm$ 2.12 & 33 $\pm$ 3 & \textbf{0.378 $\pm$ 0.055} & \textbf{0.921 $\pm$ 0.098} & \textbf{0.78 $\pm$ 0.32} & 17.88 $\pm$ 2.35 \\
& \textbf{Learnable CP} & \textbf{93.61\%} & 20.93 $\pm$ 2.03 & 31 $\pm$ 3 & 0.318 $\pm$ 0.048 & 0.816 $\pm$ 0.092 & 1.65 $\pm$ 0.58 & 16.89 $\pm$ 2.25 \\
\midrule
\multirow{3}{*}{\textbf{narrow\_graph}} 
& Naive & 71.97\% & \textbf{31.39 $\pm$ 3.25} & \textbf{66 $\pm$ 6} & 0.215 $\pm$ 0.051 & 0.441 $\pm$ 0.078 & 7.61 $\pm$ 2.12 & \textbf{34.65 $\pm$ 3.85} \\
& Standard CP & 85.18\% & 36.13 $\pm$ 3.48 & 75 $\pm$ 7 & \textbf{0.335 $\pm$ 0.061} & \textbf{0.621 $\pm$ 0.085} & \textbf{3.03 $\pm$ 1.02} & 39.88 $\pm$ 4.25 \\
& \textbf{Learnable CP} & \textbf{89.59\%} & 33.68 $\pm$ 3.31 & 69 $\pm$ 7 & 0.274 $\pm$ 0.051 & 0.525 $\pm$ 0.081 & 4.64 $\pm$ 1.38 & 37.18 $\pm$ 4.05 \\
\bottomrule
\end{tabular}%
\end{table*}

\begin{figure}[t]
\centering
\begin{minipage}{0.495\linewidth}
    \includegraphics[width=\linewidth]{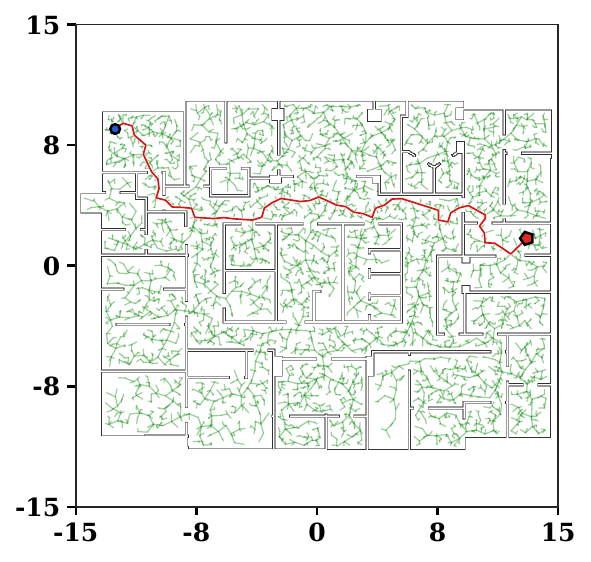}
\end{minipage}%
\begin{minipage}{0.495\linewidth}
    \includegraphics[width=\linewidth]{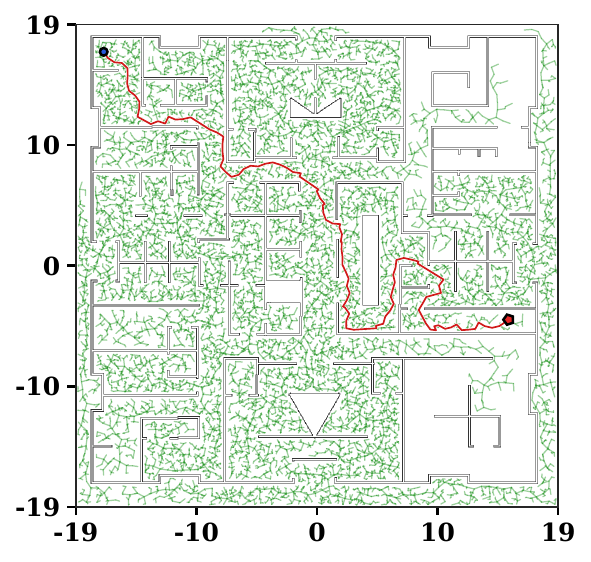}
\end{minipage}

\begin{minipage}{0.495\linewidth}
    \includegraphics[width=\linewidth]{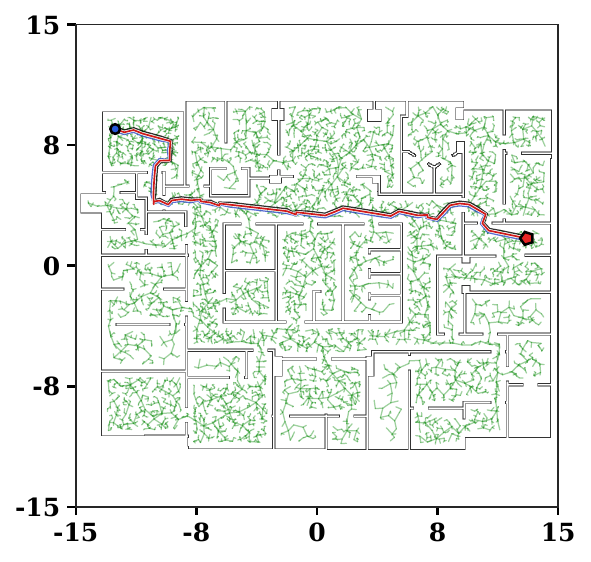}
\end{minipage}%
\begin{minipage}{0.495\linewidth}
    \includegraphics[width=\linewidth]{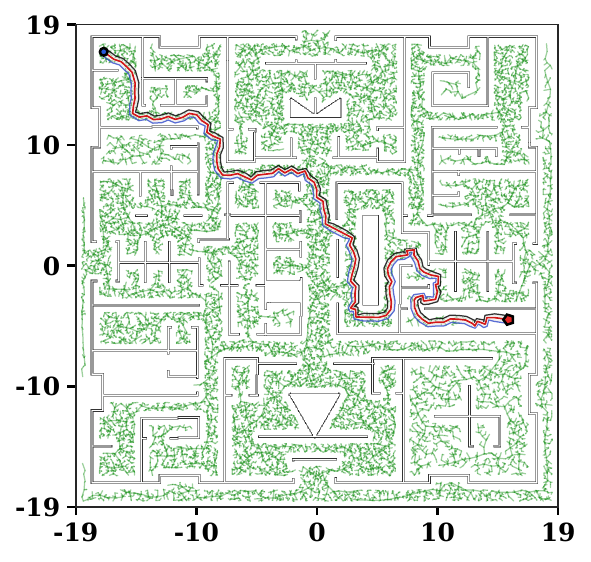}
\end{minipage}

\begin{minipage}{0.495\linewidth}
    \includegraphics[width=\linewidth]{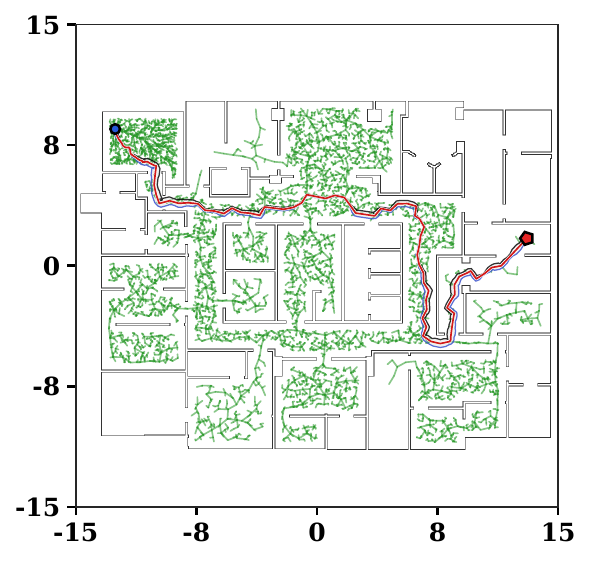}
\end{minipage}%
\begin{minipage}{0.495\linewidth}
    \includegraphics[width=\linewidth]{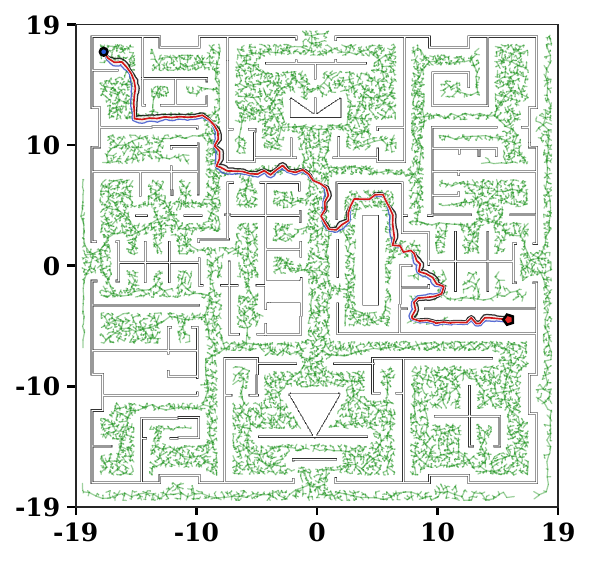}
\end{minipage}

\vspace{-4pt}
\caption{Path planning benchmarks on MRPB~\cite{amigoni2018experimental}. 
Rows correspond to methods (Naive, Standard CP, Learnable CP) and columns to environments (Office02, Shopping Mall). 
In Standard CP, red shows the naive path, while black and blue denote fixed safety margins. 
In Learnable CP, the margin adapts, wider near obstacles and tighter in open areas. 
All methods are evaluated over 1250 Monte Carlo trials using RRT*~\cite{karaman2011sampling}.}
\label{fig:path_planning}
\vspace{-10pt}
\end{figure}

\section{Robust Path Planning with Noisy and Incomplete Perception}

\subsection{Task, Environments, and Metrics}
We evaluate Learnable Conformal Prediction (LCP) on the MRPB benchmark~\cite{amigoni2018experimental} across five environments of varying complexity (some representative ones are shown in Fig.~\ref{fig:path_planning}), building on prior work in uncertainty-aware navigation \cite{darabi2024navigating}. To emulate practical constraints, we augment the benchmark with three empirically grounded sensing degradations: (i) LiDAR transparency on glass with an 18.8\% miss rate~\cite{glennie2016calibration}, (ii) partial occlusion hiding 57.5\% of obstacles~\cite{geiger2012ready}, and (iii) localization drift with 0.5\,m standard deviation~\cite{qin2018vins}. We also evaluate a combined setting including all three. Each method is tested over 1{,}250 Monte Carlo trials using RRT*~\cite{karaman2011sampling}.

We use the metrics in Table~I for characterization. Safety is measured by success rate. Path quality is assessed by path length $L$ and the number of waypoints. Safety margins are quantified by initial clearance $d_0$ (minimum obstacle distance along the planned path) and average clearance $d_{\text{avg}}$. Risk exposure is measured by danger-zone occupancy $p_0$, the fraction of the path within radius $r_0=0.20$\,m. Execution time $T$ captures efficiency. Conformal methods use a held-out calibration set disjoint from training and evaluation. Standard CP applies a single global quantile $\hat q$ across all environments and noise levels. LCP learns a parametric score $s_\theta$ and then applies the same calibration with $\hat q$ on the disjoint set.

\subsection{Results and Insights}
Table~I compares naive planning, Standard CP, and LCP across environments. LCP achieves the highest average success rate (91.5\%) while preserving near-optimal path efficiency. \textit{Evidently}, the safety-performance trade-off is inherently non-linear: increasing clearance from naive planning ($d_0 \approx 0.21$\,m) to Standard CP ($d_0 \approx 0.35$\,m) improves success by only 10\% but inflates path length by 15.3\%. In contrast, LCP’s adaptive margin ($d_0 \approx 0.29$\,m) captures nearly 90\% of the safety gain at only 40\% of the efficiency cost, showing that fixed margins fail to reflect the heterogeneous risk structure of real environments.  

Performance differences widen with environmental complexity. In structured layouts such as offices and rooms, all methods exceed 77\% success, but in narrow passages and open spaces the gap is sharper. In the shopping\_mall environment, naive planning fails in 24.4\% of trials due to obstacles, while Standard CP succeeds only by inflating path length by 9.2\%. LCP adapts its margins to context, reaching 90.4\% success with just 3.9\% additional length.  

The $p_0$ metric (danger-zone occupancy) highlights how LCP manages risk differently. Although it allows $p_0=2.9\%$ versus 1.8\% for Standard CP, it still achieves higher success by \textit{redistributing rather than uniformly minimizing risk}. LCP permits closer proximity in benign contexts (wide passages) while preserving clearance at critical points (doorways, corners). Feature attribution confirms this strategy, with passage width (score 0.033) dominating decision-making.  

Path efficiency also translates into execution-time savings. In office environments, LCP reduces average time by 8\% (19.31\,s vs. 20.98\,s), and in narrow passages by 7\% (37.18\,s vs. 39.88\,s). Over continuous operation, these savings compound. For example, a robot running 8 hours per day could complete 10–15 additional missions with LCP compared to Standard CP. Failure mode analysis further illustrates these dynamics. Naive planning achieves only 71.97\% success in narrow\_graph, reflecting its weakness in treating all uncertainties equally. Standard CP raises success to 85.18\% with conservative margins, but at the cost of longer paths (36.13\,m vs. 31.39\,m) and more waypoints (75 vs. 66), increasing cumulative exposure. LCP achieves 89.59\% success with an intermediate path length (33.68\,m), indicating that adaptive margins align more closely with actual risk distributions.

\begin{figure}[htbp]
\centering
\includegraphics[width=0.49\linewidth]{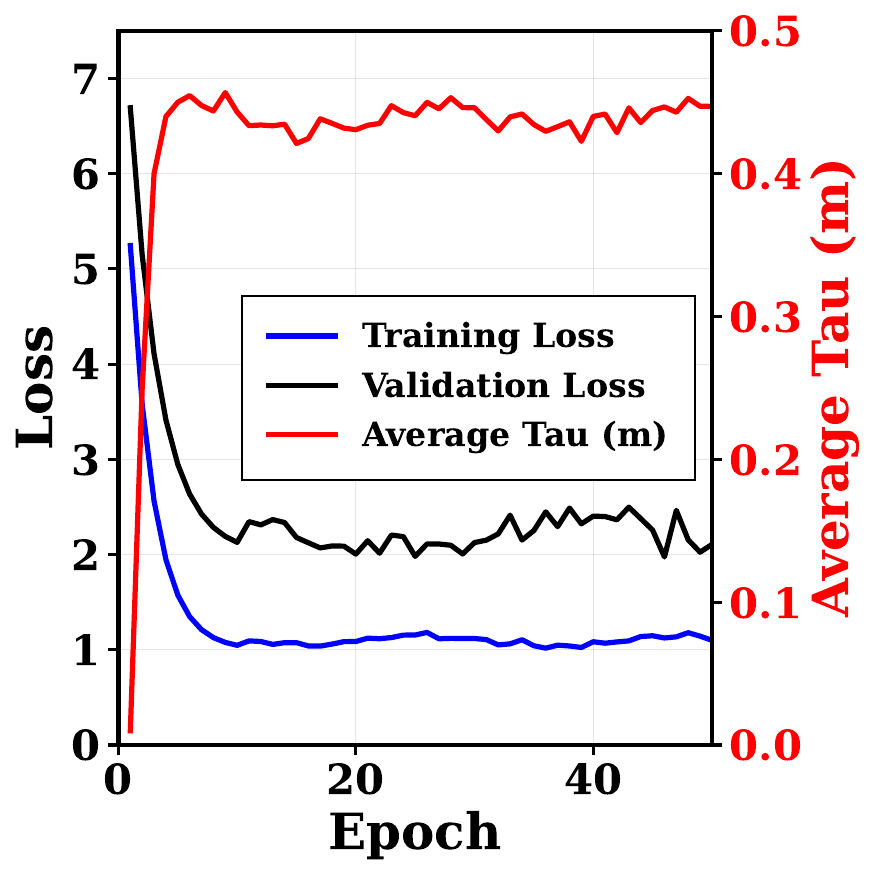}
\includegraphics[width=0.49\linewidth]{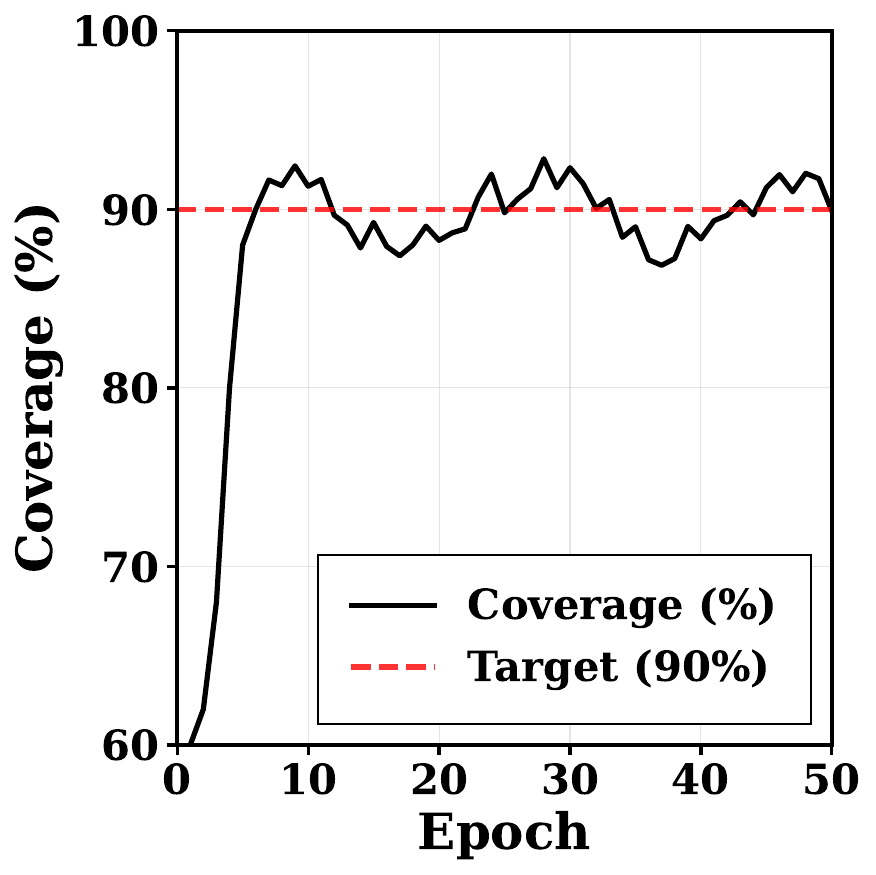}
\vspace{-20pt}\caption{Training of the scoring function over 50 epochs: \textbf{(a, left)} loss curves show fast convergence with a validation gap indicating conservative predictions, threshold $\tau$ converges to 0.44\,m via automatic calibration, and \textbf{(b, right)} coverage stabilizes at 90$\pm$3\%, confirming reliable safety ranking.}
\label{fig:training_dynamics}
\vspace{-10pt}
\end{figure}

\begin{figure}[htbp]
\centering
\includegraphics[width=0.47\columnwidth]{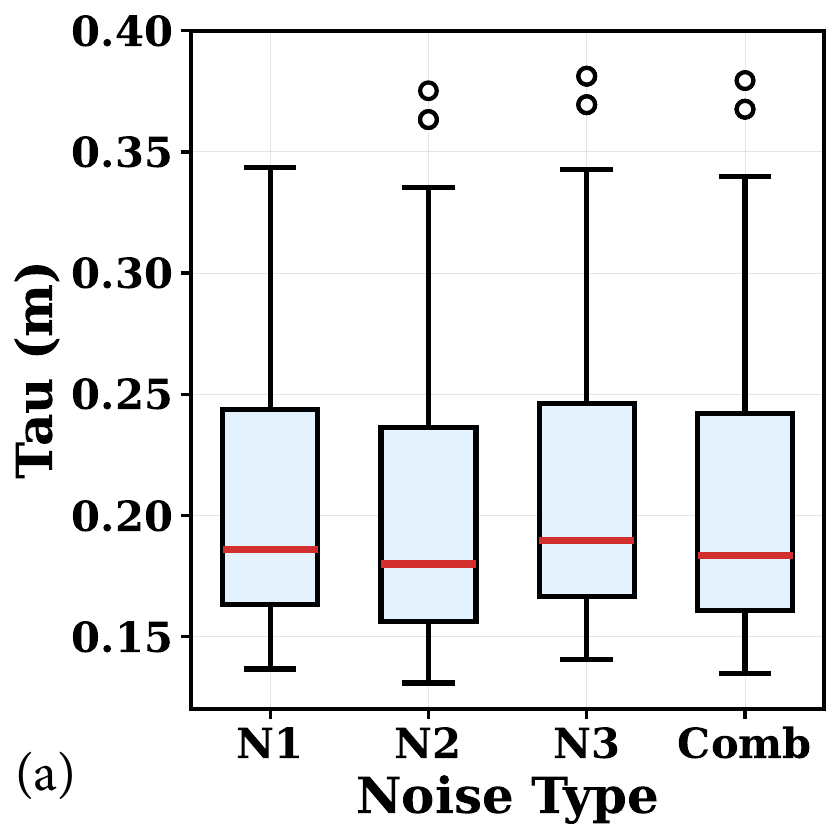}
\hfill
\includegraphics[width=0.47\columnwidth]{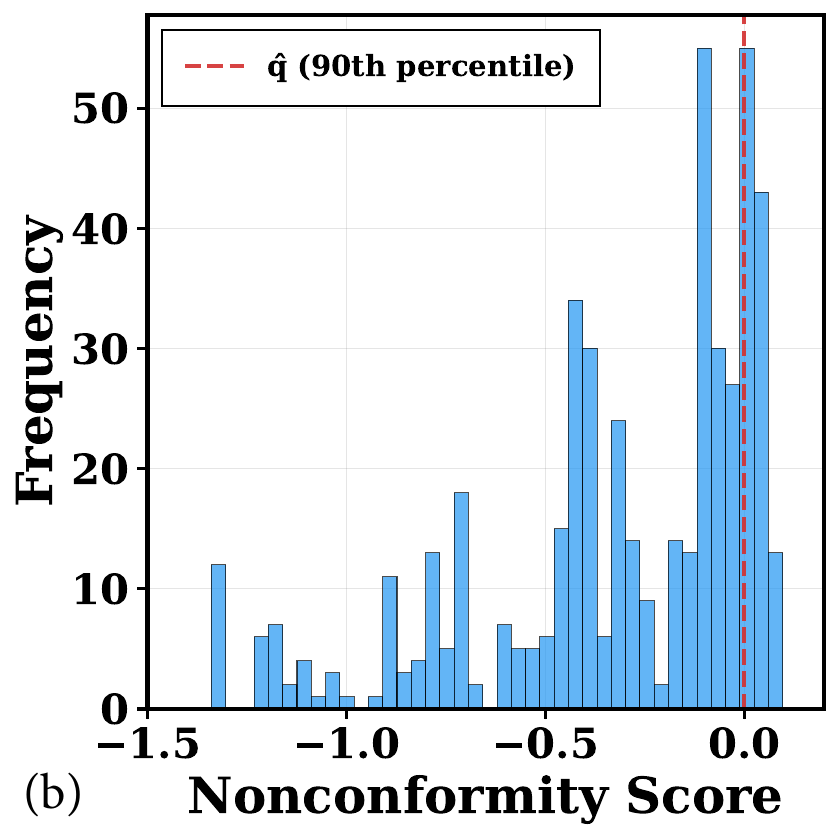}\\
\includegraphics[width=0.70\columnwidth]{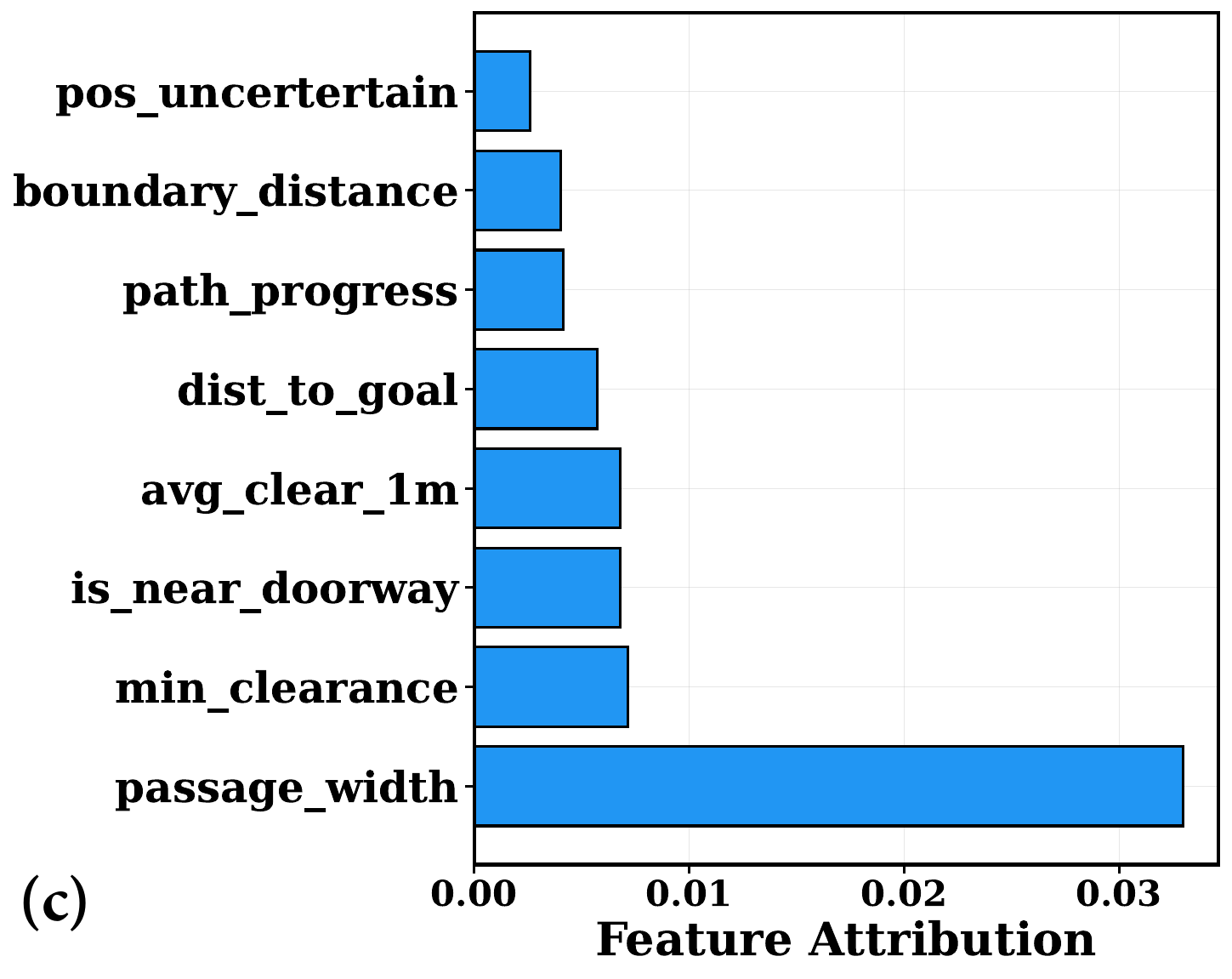}
\caption{\textbf{Ablation of Learnable CP:} \textbf{(a)} Thresholds adapt by noise, ranging 0.18–0.38 m. Here, N1 = transparency noise, N2 = occlusion noise, N3 = localization noise, and Comb = combined noise. \textbf{(b)} Bimodal score distributions reveal the gap between learned rankings and safety. \textbf{(c)} Feature importance shows geometry dominance, with passage width as key driver.}
\label{fig:ablation_analysis}
\vspace{-15pt}
\end{figure}

\subsection{Learning Dynamics and Safety Calibration}
Fig.~\ref{fig:training_dynamics} tracks LCP training over 50 epochs. Training loss drops from 5.3 to 1.1 within 10 epochs, while validation stabilizes near 2.1, leaving a persistent $\sim$1.0 gap that reflects stochasticity and indicates that general patterns are learned but conservative predictions persist on unseen data. The conformal threshold $\tau$ rises from near zero to 0.45\,m by epoch 7 and oscillates around 0.44\,m ($\pm$0.02\,m), about 38\% above Standard CP’s fixed 0.32\,m. These oscillations represent exploration of the safety–efficiency boundary rather than premature convergence. Coverage improves from 60\% to 92\% within 5 epochs and stabilizes at 90$\pm$3\%, even though deployment $\tau$ averages only 0.20\,m. This shows that LCP learns relative safety rankings rather than absolute thresholds, with coverage maintained through post-hoc calibration. 

\begin{figure}[htbp]
\centering
\includegraphics[width=\columnwidth]{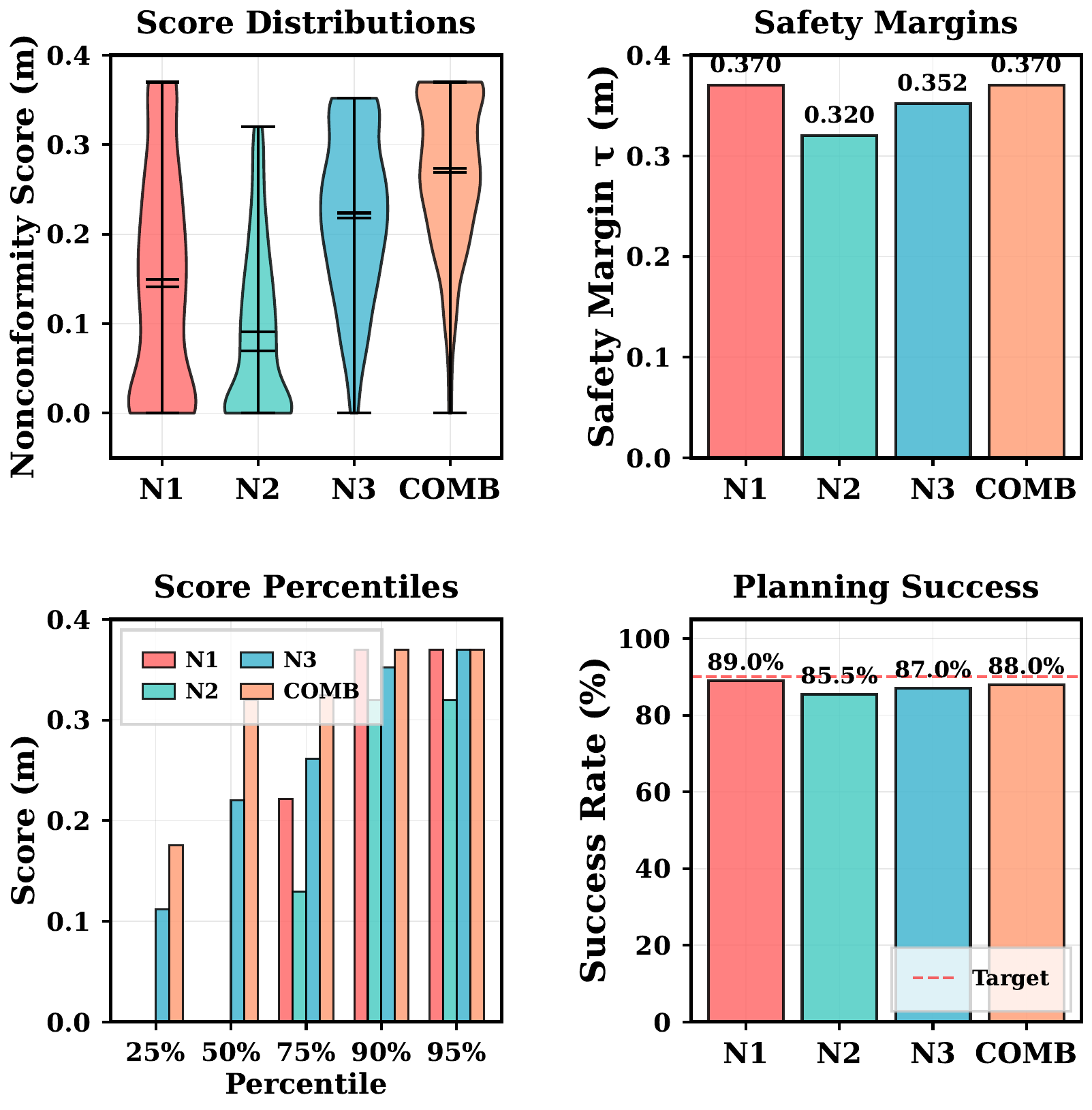}
\vspace{-20pt}\caption{Standard CP ablation across noise types: Here, N1 = transparency noise, N2 = occlusion noise, N3 = localization noise, and Comb = combined noise. Occlusion yields tight score distributions, transparency shows wider spread, margins adjust from 0.32--0.37\,m, calibration holds at the 90th percentile, and success rates cluster at 85.5--89.0\%.}
\label{fig:std_cp_ablation}
\end{figure}

Ablation studies (Fig.~\ref{fig:ablation_analysis}) show LCP achieves context-aware safety through adaptive thresholds: occlusion yields narrow margins (0.16--0.24\,m), localization spans 0.14--0.34\,m, with crisis-response outliers at 0.36--0.38\,m. Bimodal score distributions peak at -0.4\,m (efficiency) and -0.05\,m (safety), confirming ordinal safety learning across 0.13--0.38\,m margins. Feature attribution identifies passage width (0.033) as the key driver, while minimum clearance contributes little. Negative density/occlusion attributions indicate margins shrink in clutter to preserve navigability.  

\begin{table*}[!t]
    \centering
    \textbf{Table II: Uncertainty Quantification for Object Localization across Models and Datasets (target: 0.90).} \vspace{2pt}
    \label{tab:main_results}
    \setlength{\tabcolsep}{2pt} 
    \begin{tabular}{llcccccc}
    \toprule
    & & \multicolumn{2}{c}{\textbf{COCO}} & \multicolumn{2}{c}{\textbf{BDD100K}} & \multicolumn{2}{c}{\textbf{Cityscapes}} \\
    \cmidrule(lr){3-4} \cmidrule(lr){5-6} \cmidrule(lr){7-8}
    \textbf{Base Model} & \textbf{Method} & Cov. & MPIW & Cov. & MPIW & Cov. & MPIW \\
    \midrule
    \multirow{4}{*}{ResNeXt-101-FPN*} 
    & Standard CP & \SI{0.900 \pm 0.013}{} & \SI{90.6 \pm 9.3}{} & \SI{0.919 \pm 0.009}{} & \SI{59.8 \pm 3.3}{} & \SI{0.912 \pm 0.029}{} & \SI{100.0 \pm 20.3}{} \\
    & Ensemble & \SI{0.927 \pm 0.005}{} & \SI{109.7 \pm 3.7}{} & \SI{0.900 \pm 0.037}{} & \SI{80.4 \pm 7.1}{} & \SI{0.906 \pm 0.037}{} & \SI{127.6 \pm 16.1}{} \\
    & CQR & \SI{0.891 \pm 0.010}{} & \SI{87.7 \pm 13.6}{} & \SI{0.910 \pm 0.006}{} & \SI{71.0 \pm 4.4}{} & \SI{0.908 \pm 0.063}{} & \SI{110.0 \pm 25.9}{} \\
    & \textbf{Learnable (Ours)} & \textbf{\SI{0.902 \pm 0.020}{}} & \textbf{\SI{41.9 \pm 1.8}{}} & \textbf{\SI{0.896 \pm 0.019}{}} & \textbf{\SI{28.8 \pm 1.2}{}} & \textbf{\SI{0.887 \pm 0.021}{}} & \textbf{\SI{53.8 \pm 2.1}{}} \\
    \midrule
    \multirow{2}{*}{Cascade R-CNN} 
    & Standard CP & \SI{0.927 \pm 0.008}{} & \SI{109.0 \pm 9.6}{} & \SI{0.912 \pm 0.025}{} & \SI{54.3 \pm 2.3}{} & \SI{0.924 \pm 0.065}{} & \SI{96.6 \pm 12.3}{} \\
    & \textbf{Learnable (Ours)} & \textbf{\SI{0.898 \pm 0.018}{}} & \textbf{\SI{37.3 \pm 1.5}{}} & \textbf{\SI{0.892 \pm 0.017}{}} & \textbf{\SI{27.5 \pm 1.1}{}} & \textbf{\SI{0.903 \pm 0.019}{}} & \textbf{\SI{51.2 \pm 2.0}{}} \\
    \midrule
    \multirow{2}{*}{ResNet-50-FPN} 
    & Standard CP & \SI{0.900 \pm 0.011}{} & \SI{105.9 \pm 11.3}{} & \SI{0.901 \pm 0.013}{} & \SI{58.7 \pm 3.6}{} & \SI{0.900 \pm 0.022}{} & \SI{95.5 \pm 18.7}{} \\
    & \textbf{Learnable (Ours)} & \textbf{\SI{0.891 \pm 0.020}{}} & \textbf{\SI{46.4 \pm 1.6}{}} & \textbf{\SI{0.885 \pm 0.018}{}} & \textbf{\SI{29.1 \pm 1.3}{}} & \textbf{\SI{0.896 \pm 0.022}{}} & \textbf{\SI{57.9 \pm 2.2}{}} \\
    \midrule
    \multirow{2}{*}{ResNet-50-C4} 
    & Standard CP & \SI{0.939 \pm 0.016}{} & \SI{120.7 \pm 10.6}{} & \SI{0.910 \pm 0.048}{} & \SI{62.8 \pm 4.6}{} & \SI{0.910 \pm 0.038}{} & \SI{128.3 \pm 35.0}{} \\
    & \textbf{Learnable (Ours)} & \textbf{\SI{0.905 \pm 0.021}{}} & \textbf{\SI{51.2 \pm 1.9}{}} & \textbf{\SI{0.900 \pm 0.019}{}} & \textbf{\SI{32.2 \pm 1.4}{}} & \textbf{\SI{0.910 \pm 0.023}{}} & \textbf{\SI{62.4 \pm 2.4}{}} \\
    \midrule
    \multirow{2}{*}{ResNeXt-INT8} 
    & Standard CP & \SI{0.885 \pm 0.015}{} & \SI{95.2 \pm 10.8}{} & \SI{0.895 \pm 0.020}{} & \SI{65.5 \pm 5.2}{} & \SI{0.900 \pm 0.025}{} & \SI{115.8 \pm 22.1}{} \\
    & \textbf{Learnable (Ours)} & \textbf{\SI{0.894 \pm 0.019}{}} & \textbf{\SI{55.8 \pm 2.1}{}} & \textbf{\SI{0.888 \pm 0.018}{}} & \textbf{\SI{35.7 \pm 1.5}{}} & \textbf{\SI{0.899 \pm 0.020}{}} & \textbf{\SI{66.7 \pm 2.6}{}} \\
    \bottomrule
    \end{tabular}
\end{table*}

LCP's multi-scale reasoning uses geometry for baselines, uncertainty for bounds, and nonlinear mappings for context. Standard CP (Fig.~\ref{fig:std_cp_ablation}) adapts coarsely: $\tau$ shifts 0.32--0.37\,m between noise types, with success clustering at 85.5--89.0\% from discrete quantization. While recognizing uncertainty patterns, it lacks LCP's continuous adaptation that achieves 91.5\% success with near-optimal efficiency. The scoring network uses residual connections with batch normalization and proper weight initialization to ensure stable convergence across environments.

\FloatBarrier

\section{Perception Robustness under Uncertainty}

\subsection{Task, Datasets, and Setup}
We evaluate uncertainty-aware object detection on COCO~\cite{lin2014microsoft}, BDD100K~\cite{yu2020bdd100k}, and Cityscapes~\cite{cordts2016cityscapes}, comparing LCP with Standard CP, deep ensembles, and conformalized quantile regression (CQR)~\cite{timans2024adaptive}. 

Our hardware evaluation on Intel NUC (Core Ultra 7 165H, 64\,GB DDR5, Intel Arc GPU, AI Boost NPU) shows that LCP adds $<$1\% memory overhead and 15.9\% inference overhead ($\sim$3.5\,ms per frame), while maintaining 39\,FPS for YOLOv8n object detection. Compared to Standard CP, LCP halves interval width (102\,px $\rightarrow$ 51\,px) at the same 90\% coverage, with modest power increase (35\,W $\rightarrow$ 38\,W).

For object detection, the feature extraction pipeline processes each bounding box through a carefully designed 13-dimensional encoding: normalized coordinates $(x_0, y_0, x_1, y_1)$, detection confidence $c$, and geometric descriptors including $\log(\text{area})$, aspect ratio, centroid position relative to image center, and distance to image boundaries. These features capture both spatial uncertainty (objects near edges are harder to localize) and scale-dependent difficulty (small objects require tighter tolerances). 


\subsection{Results and Insights}
\begin{figure*}[!t]
\centering
\includegraphics[width=0.9\textwidth]{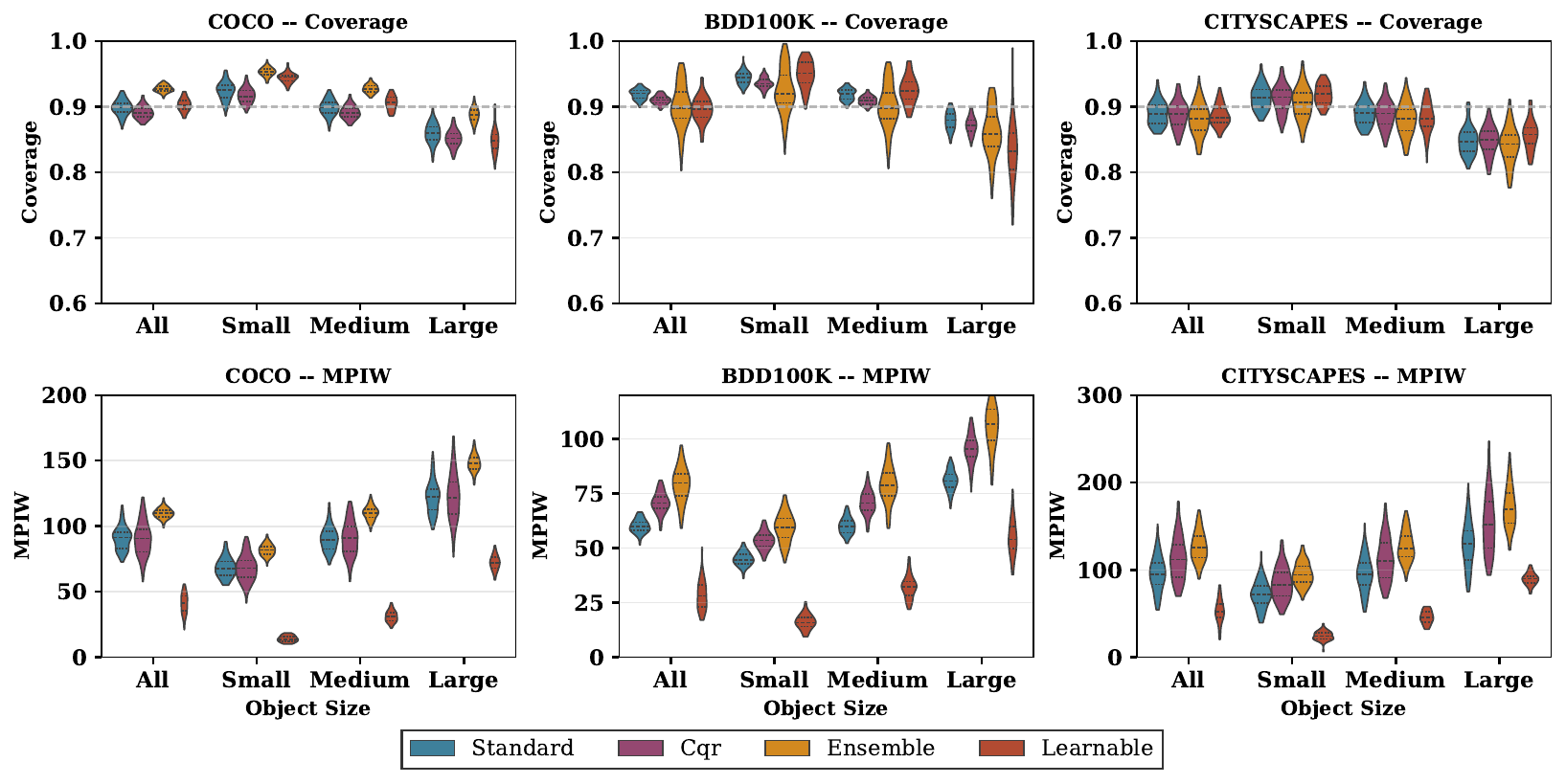}
\vspace{-10pt}\caption{Violin plots of coverage and MPIW on COCO, BDD100K, and Cityscapes. Our Learnable CP (red) achieves the tightest intervals with coverage near 90\%, reducing small-object MPIW to ~15 pixels (vs. ~70) and large-object MPIW to ~70 pixels (vs. 120–150), demonstrating superior coverage-efficiency.}\vspace{-10pt}
\label{fig:violin_plots}
\end{figure*}

Our evaluations on object detection and classification highlight several key properties of LCP. First, it achieves scale-aware efficiency gains by sharply tightening prediction intervals while preserving validity. As Table~II and Fig.~\ref{fig:violin_plots} show, average MPIW contracts to 41.9 px on COCO, 28.8 px on BDD100K, and 53.8 px on Cityscapes, which is 46--54\% smaller than Standard CP while maintaining coverage near 90\%. Crucially, this improvement is not uniform compression but adaptive redistribution of margins. In Fig.~\ref{fig:violin_plots}, small objects, often under-protected by fixed thresholds, reach 94.2\% coverage with only 14.7 px slack, whereas large objects, which otherwise inflate margins excessively, settle at 84.7\% coverage with 72.2 px slack. These scale-specific tradeoffs show how LCP encodes task difficulty directly into interval allocation, offering higher protection where risk is greatest and greater efficiency where inflation dominates.

\begin{figure}[t]
\centering
\includegraphics[trim=10 10 0 0, clip, width=0.48\textwidth]{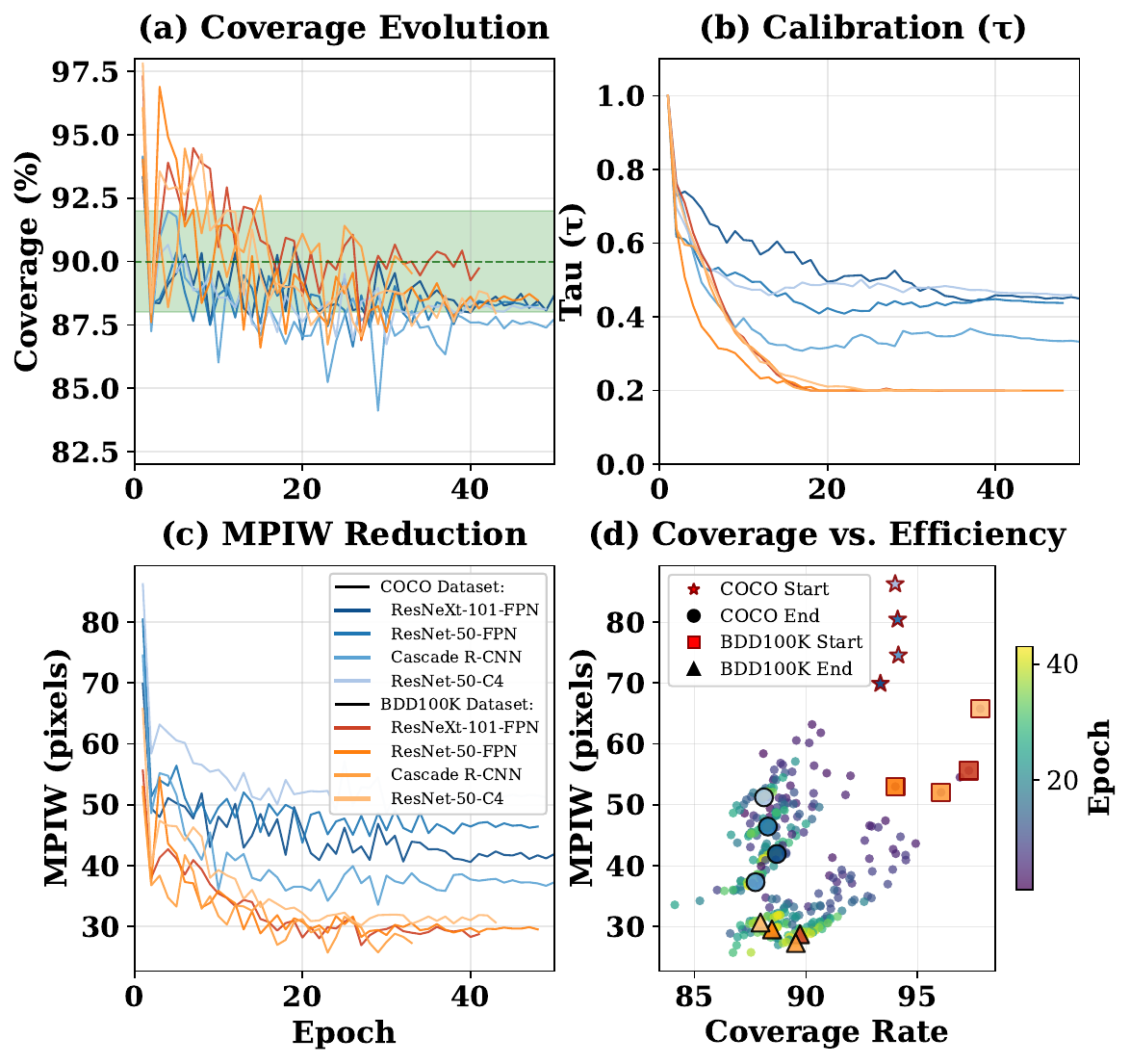}
\vspace{-20pt}\caption{Multi-architecture $\tau$ calibration analysis. \textbf{(a)} Coverage stabilizes in the 88--92\% zone for COCO (blue) and BDD100K (orange). \textbf{(b)} $\tau$ adapts from 1.0 to dataset-specific optima ($\approx$0.45, 0.2). \textbf{(c)} MPIW shrinks 30--50\% (80$\rightarrow$46 px). \textbf{(d)} Coverage-efficiency trajectories converge to Pareto-optimal fronts within 20--30 epochs, consistent across detector backbones.}\vspace{-10pt}
\label{fig:tau_calibration}
\end{figure}

\begin{figure}[t]
\centering
\includegraphics[trim=5 10 0 0, clip, width=0.48\textwidth]{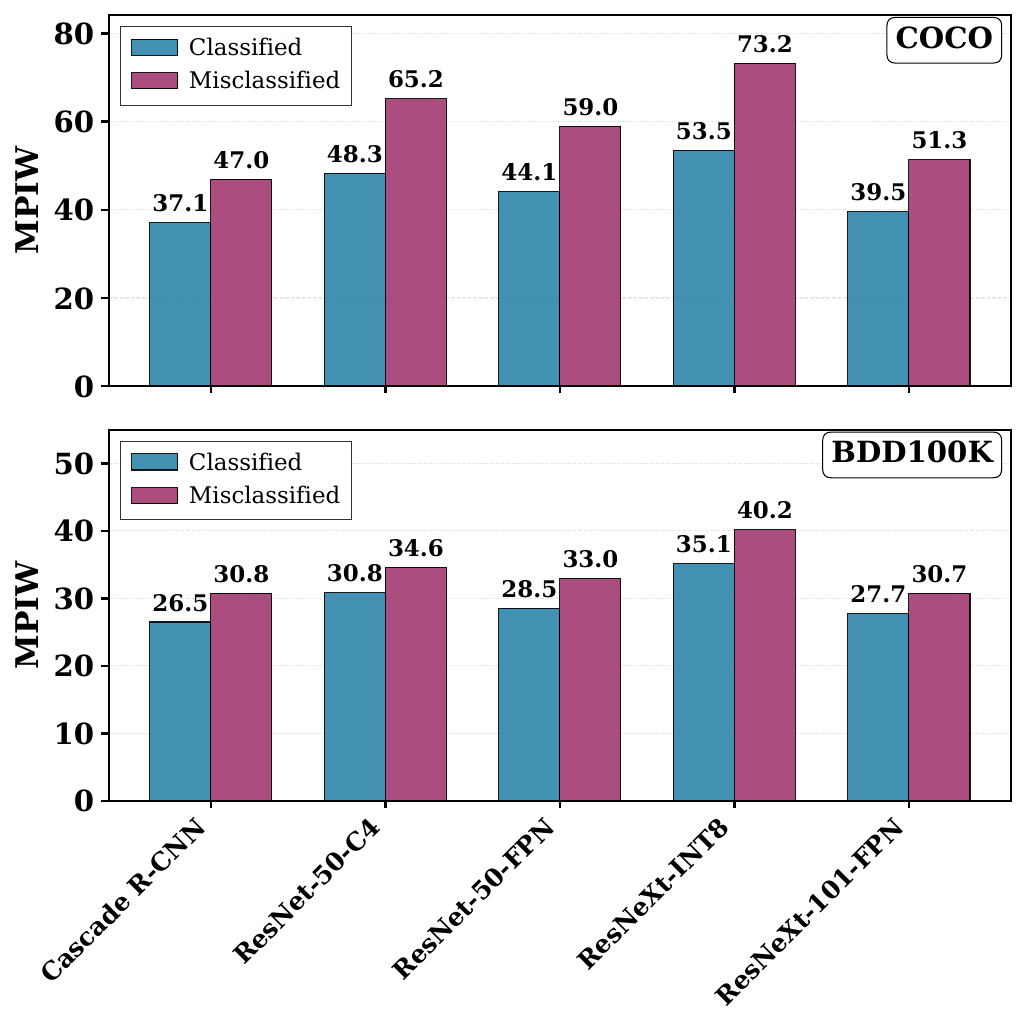}
\vspace{-15pt}\caption{MPIW comparison on COCO and BDD100K showing misclassified predictions receive wider intervals, confirming adaptive uncertainty quantification across architectures.}\vspace{-10pt}
\label{fig:misclassification}
\end{figure}

The method also demonstrates calibration stability across backbones. As shown in Fig.~\ref{fig:tau_calibration}, coverage quickly stabilizes in the 88--92\% band, $\tau$ converges to dataset-specific optima (0.45 for COCO, 0.20 for BDD100K), and MPIW shrinks by 30--50\% within 20--30 epochs. Coverage--efficiency trajectories reveal that LCP begins in inefficient, high-MPIW regimes but steadily tracks toward Pareto-optimal fronts. This pattern repeats across architectures including ResNeXt-101, Cascade R-CNN, ResNet-50-FPN/C4, and quantized ResNeXt-INT8, confirming that learned nonconformity functions transfer across detectors without the repeated calibration sweeps static scores require. In effect, uncertainty calibration becomes a one-shot training problem rather than an ongoing maintenance overhead.

The training process follows a phased approach to balance multiple objectives. Initially, the network learns to discriminate between correct and incorrect predictions through a margin loss. Subsequently, coverage constraints are introduced via quadratic penalties when empirical coverage deviates from targets. Finally, efficiency optimization minimizes the Mean Prediction Interval Width (MPIW) normalized by object size, preventing trivial solutions that achieve coverage through excessive inflation.

Another distinctive behavior of LCP is its ability to widen intervals when predictions are unreliable. As Fig.~\ref{fig:misclassification} shows, misclassified detections receive 33\% wider intervals on COCO and 14\% on BDD100K compared to correct ones. This asymmetry reflects calibrated caution rather than random noise, similar to the principled abstention mechanisms explored in \cite{tayebati2025learningconformal}. By linking interval inflation to prediction difficulty, LCP communicates uncertainty in a way downstream planners can exploit. In autonomous driving, for instance, margins expand only when detections are dubious, allowing braking or re-planning to be applied selectively rather than conservatively at all times. Baselines, by contrast, inflate uniformly and cannot convey this risk.

\begin{figure}[t]
\centering
\includegraphics[width=\columnwidth]{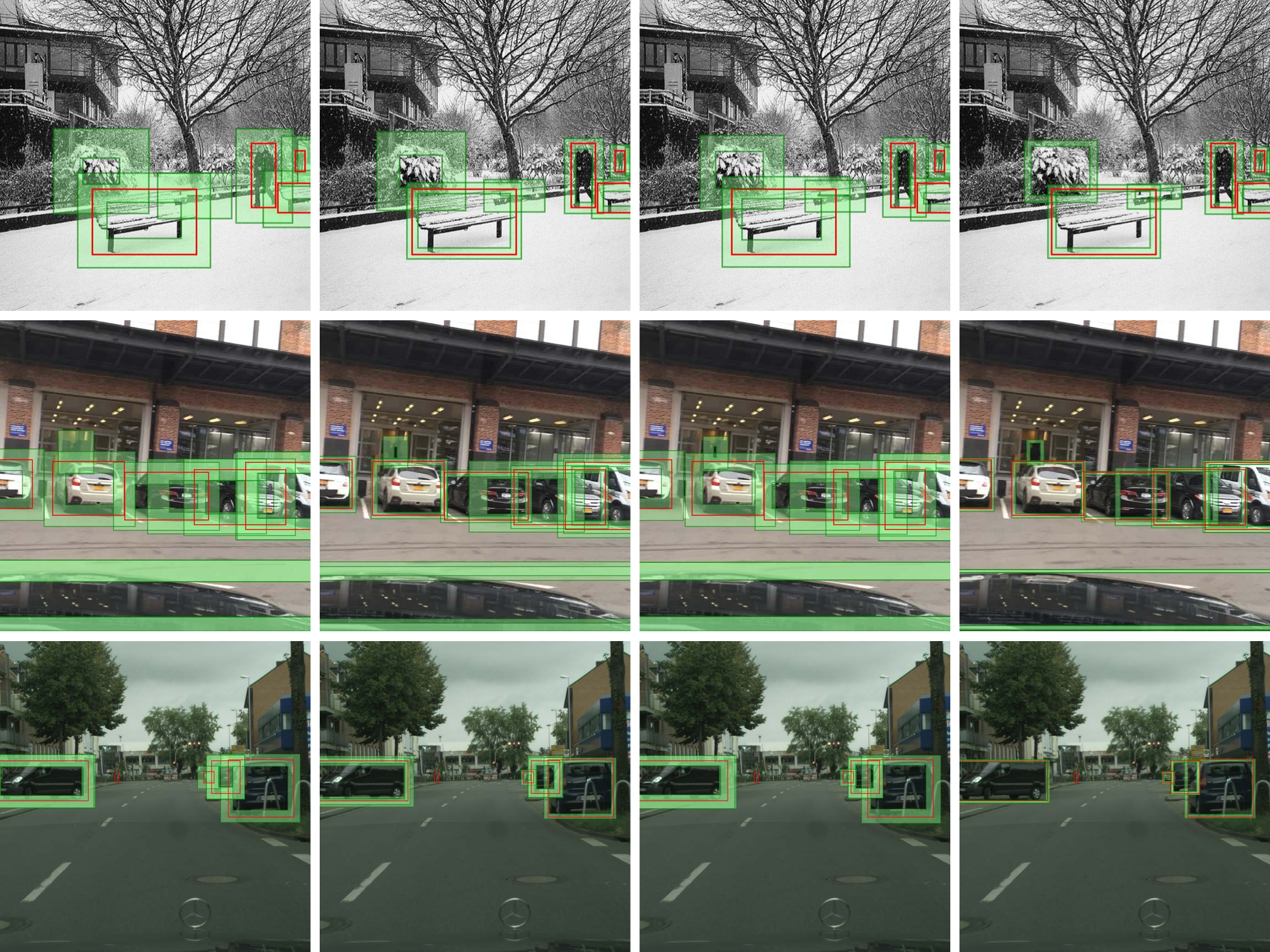}
\vspace{-10pt}\caption{Examples of conformal prediction intervals produced by different scoring functions across three datasets. Rows from top to bottom: COCO, BDD100K, and Cityscapes. Columns from left to right: Standard, Ensemble, CQR, and Learnable CP. True bounding boxes are in red, prediction boxes are in green, and prediction interval regions are shaded in green.}
\label{fig:bounding_boxes}
\vspace{-10pt}
\end{figure}

Comparatively, standard CP applies rigid margins that ignore context, ensembles inflate excessively (often doubling box area), and CQR produces irregular, geometrically implausible shapes. LCP instead yields tight, well-formed intervals that scale with object size and difficulty. On COCO, small-object intervals shrink from $\sim$70 px in baselines to 15 px, while large-object intervals contract from 120--150 px to $\sim$70 px. In BDD100K, dominated by occlusion and lighting variation, LCP halves interval width while sustaining $\approx$90\% coverage. In Cityscapes, with dense urban clutter, it avoids the excessive inflation of ensembles while still preserving coverage. These results emphasize not only efficiency but also interpretability: LCP produces compact, symmetric, and stable intervals, making uncertainty actionable. Fig.~\ref{fig:bounding_boxes} visualizes these differences across the three datasets, showing how LCP's adaptive intervals tightly bound objects while standard methods either over-inflate uniformly or produce irregular shapes.

The calibration process employs smoothed quantile estimation with asymmetric windows to ensure conservative yet stable thresholds. Rather than using a single empirical quantile, we compute a weighted average over a 5-sample window with weights $w_i \in [1.0, 1.5]$ that emphasize samples below the target quantile. This design choice prevents threshold oscillation while maintaining the finite-sample coverage guarantee required by conformal prediction theory.

Finally, the classification results in Table~III confirm that the benefits of learned scoring generalize beyond detection. Across CIFAR-100, HAM10000, ImageNet, Places365, and PlantNet, LCP consistently yields the smallest prediction sets, shrinking average size by 4.7--9.9\% relative to fixed baselines while maintaining competitive coverage. In challenging domains such as PlantNet, where baselines degenerate to trivial AUROC ($\approx$0.71), LCP recovers discriminative power (AUROC $\approx$0.99) with far leaner sets. These findings show that learned scoring preserves coverage while improving both efficiency and calibration across recognition tasks.

  \begin{table}[!t]
  \setlength{\tabcolsep}{3pt}
  \renewcommand{\arraystretch}{0.9}
  \centering
  \textbf{Table III: Learnable vs. static scoring on classification.}\vspace{2pt}
  \label{tab:classification_results}
  \begin{tabular}{l l cccc}
  \toprule
  \textbf{} & \textbf{Method} & \textbf{Coverage $\uparrow$} & \textbf{Set Size $\downarrow$} & \textbf{AUROC $\uparrow$} & \textbf{ECE $\downarrow$} \\
  \midrule
  \multirow{5}{*}{\rotatebox{90}{\textbf{CIFAR-100}}}
      & 1-p (Baseline)      & \num{89.64} & \num{1.73} & \num{0.97} & \num{0.04} \\
      & APS                 & \textbf{\num{90.38}} & \num{3.71} & \num{0.95} & \num{0.11} \\
      & LogMargin           & \num{90.14} & \num{2.47} & \num{0.96} & \num{0.06} \\
      & Sparsemax           & \num{88.44} & \num{2.21} & \num{0.92} & \num{0.03} \\
      & \textbf{Ours}       & \num{88.01} & \textbf{\num{1.56}} & \textbf{\num{0.98}} & \textbf{\num{0.02}} \\
  \midrule
  \multirow{5}{*}{\rotatebox{90}{\textbf{HAM10000}}}
      & 1-p (Baseline)      & \num{90.24} & \num{1.31} & \num{0.96} & \num{0.04} \\
      & APS                 & \textbf{\num{90.34}} & \num{1.25} & \num{0.94} & \num{0.05} \\
      & LogMargin           & \num{89.94} & \num{1.51} & \num{0.94} & \num{0.03} \\
      & Sparsemax           & \num{90.04} & \num{1.12} & \num{0.88} & \num{0.02} \\
      & \textbf{Ours}       & \num{89.31} & \textbf{\num{1.06}} & \textbf{\num{0.97}} & \textbf{\num{0.01}} \\
  \midrule
  \multirow{5}{*}{\rotatebox{90}{\textbf{ImageNet}}}
      & 1-p (Baseline)      & \num{90.15} & \num{1.53} & \num{0.97} & \num{0.02} \\
      & APS                 & \num{90.16} & \num{2.05} & \num{0.95} & \num{0.04} \\
      & LogMargin           & \num{90.06} & \num{1.66} & \num{0.96} & \num{0.02} \\
      & Sparsemax           & \textbf{\num{90.55}} & \num{1.80} & \num{0.93} & \num{0.01} \\
      & \textbf{Ours}       & \num{89.57} & \textbf{\num{1.46}} & \textbf{\num{0.98}} & \textbf{\num{0.01}} \\
  \midrule
  \multirow{5}{*}{\rotatebox{90}{\textbf{Places365}}}
      & 1-p (Baseline)      & \num{90.12} & \num{17.99} & \num{0.95} & \num{0.02} \\
      & APS                 & \num{90.10} & \num{21.06} & \num{0.96} & \num{0.04} \\
      & LogMargin           & \num{90.16} & \num{16.64} & \num{0.97} & \num{0.03} \\
      & Sparsemax           & \textbf{\num{90.59}} & \num{22.55} & \num{0.94} & \num{0.06} \\
      & \textbf{Ours}       & \num{89.85} & \textbf{\num{14.97}} & \textbf{\num{0.97}} & \textbf{\num{0.01}} \\
  \midrule
  \multirow{5}{*}{\rotatebox{90}{\textbf{PlantNet}}}
      & 1-p (Baseline)      & \num{89.66} & \num{10.08} & \num{0.71} & \num{0.02} \\
      & APS                 & \num{89.72} & \num{13.27} & \num{0.70} & \num{0.03} \\
      & LogMargin           & \num{89.42} & \num{12.36} & \num{0.78} & \textbf{\num{0.01}} \\
      & Sparsemax           & \textbf{\num{90.93}} & \num{12.42} & \num{0.79} & \num{0.14} \\
      & \textbf{Ours}       & \num{89.08} & \textbf{\num{6.73}} & \textbf{\num{0.94}} & \num{0.02} \\
  \bottomrule
  \end{tabular}%
  \vspace{-8pt}
  \end{table}

\section{Conclusion}
We introduced Learnable Conformal Prediction (LCP), a framework that learns context-aware nonconformity functions while preserving the guarantees of conformal prediction. On the MRPB benchmark, LCP raised navigation success to 91.5\% with only 4.5\% path inflation, compared to 12.2\% for Standard CP. For object detection on COCO, BDD100K, and Cityscapes, it reduced interval widths by 46--54\% at 90\% coverage, and in classification on HAM10000 and ImageNet it cut set sizes by 4.7--9.9\%. The method is lightweight ($\sim$4.8\% runtime overhead, 42\,KB memory) and supports online adaptation, making it well suited to resource-constrained autonomous systems. 
Hardware evaluation on Intel NUC (Core Ultra 7 165H) shows that LCP adds $<$1\% memory and 15.9\% inference overhead ($\sim$3.5\,ms), yet sustains 39\,FPS on YOLOv8n detection. LCP requires only 0.97 mJ per frame versus 7.22 mJ for ensembles—7.4× more energy-efficient—with INT8 quantization enabling 78 FPS operation. At equal 90\% coverage, it halves interval width (102\,px $\rightarrow$ 51\,px) with only a modest power rise (35\,W $\rightarrow$ 38\,W).\\
\noindent\textbf{Acknowledgement:}
This work was supported by CogniSense, one of the several SR/DARPA JUMP2.0 Centers, NSF Grant\# 2046435, and UI-UAAT Collaboration Funds.

\bibliographystyle{IEEEtran}
\bibliography{references}

\begin{thebibliography}{10}
\providecommand{\url}[1]{#1}
\csname url@samestyle\endcsname
\providecommand{\newblock}{\relax}
\providecommand{\bibinfo}[2]{#2}
\providecommand{\BIBentrySTDinterwordspacing}{\spaceskip=0pt\relax}
\providecommand{\BIBentryALTinterwordstretchfactor}{4}
\providecommand{\BIBentryALTinterwordspacing}{\spaceskip=\fontdimen2\font plus
\BIBentryALTinterwordstretchfactor\fontdimen3\font minus \fontdimen4\font\relax}
\providecommand{\BIBforeignlanguage}[2]{{%
\expandafter\ifx\csname l@#1\endcsname\relax
\typeout{** WARNING: IEEEtran.bst: No hyphenation pattern has been}%
\typeout{** loaded for the language `#1'. Using the pattern for}%
\typeout{** the default language instead.}%
\else
\language=\csname l@#1\endcsname
\fi
#2}}
\providecommand{\BIBdecl}{\relax}
\BIBdecl

\bibitem{guo2017calibration}
C.~Guo, G.~Pleiss, Y.~Sun, and K.~Q. Weinberger, ``On calibration of modern neural networks,'' in \emph{Proceedings of the 34th International Conference on Machine Learning}, 2017, pp. 1321--1330.

\bibitem{taori2020shift}
R.~Taori, A.~Dave, V.~Shankar, N.~Carlini, B.~Recht, and L.~Schmidt, ``Measuring robustness to natural distribution shifts in image classification,'' in \emph{Advances in Neural Information Processing Systems}, vol.~33, 2020, pp. 18\,583--18\,599.

\bibitem{amodei2016concrete}
D.~Amodei, C.~Olah, J.~Steinhardt, P.~Christiano, J.~Schulman, and D.~Man{\'e}, ``Concrete problems in ai safety,'' \emph{arXiv preprint arXiv:1606.06565}, 2016.

\bibitem{der2009aleatoric}
A.~Der~Kiureghian and O.~Ditlevsen, ``Aleatoric and epistemic uncertainty in machine learning: An introduction to concepts and methods,'' \emph{Journal of Reliability Engineering \& System Safety}, vol.~94, no.~3, pp. 105--109, 2009.

\bibitem{kendall2017uncertainties}
A.~Kendall and Y.~Gal, ``What uncertainties do we need in bayesian deep learning for computer vision?'' in \emph{Advances in Neural Information Processing Systems}, 2017, pp. 5574--5584.

\bibitem{zhang2024conformal}
L.~Zhang, R.~Kamath, S.~Wang, E.~Schmerling, and M.~Pavone, ``Conformal prediction for safe autonomous driving under distribution shifts,'' in \emph{2024 Conference on Robot Learning (CoRL)}.\hskip 1em plus 0.5em minus 0.4em\relax PMLR, 2024, pp. 1842--1851.

\bibitem{luo2024efficient}
R.~Luo, S.~Zhao, J.~Koenig, S.~Adeli, and M.~Pavone, ``Efficient uncertainty quantification for vision-based autonomous navigation,'' in \emph{2024 IEEE/RSJ International Conference on Intelligent Robots and Systems (IROS)}.\hskip 1em plus 0.5em minus 0.4em\relax IEEE, 2024, pp. 8234--8241.

\bibitem{park2024adaptive}
H.~Park, M.~Meghjani, and M.~H.~A. Lim, ``Adaptive conformal prediction for human-robot collaboration under covariate shift,'' \emph{IEEE Robotics and Automation Letters}, vol.~9, no.~3, pp. 2456--2463, 2024.

\bibitem{stutts2024conformal}
A.~C. Stutts, D.~Kumar, T.~Tulabandhula, and A.~Trivedi, ``Invited: Conformal inference meets evidential learning: Distribution-free uncertainty quantification with epistemic and aleatoric separability,'' in \emph{Proceedings of the 61st ACM/IEEE Design Automation Conference}, ser. DAC '24.\hskip 1em plus 0.5em minus 0.4em\relax New York, NY, USA: Association for Computing Machinery, 2024.

\bibitem{blundell2015weight}
C.~Blundell, J.~Cornebise, K.~Kavukcuoglu, and D.~Wierstra, ``Weight uncertainty in neural networks,'' in \emph{Proceedings of the 32nd International Conference on Machine Learning}, 2015, pp. 1613--1622.

\bibitem{lakshminarayanan2017simple}
B.~Lakshminarayanan, A.~Pritzel, and C.~Blundell, ``Simple and scalable predictive uncertainty estimation using deep ensembles,'' in \emph{Advances in Neural Information Processing Systems}, 2017.

\bibitem{gal2016dropout}
Y.~Gal and Z.~Ghahramani, ``Dropout as a bayesian approximation: Representing model uncertainty in deep learning,'' in \emph{International Conference on Machine Learning}, 2016, pp. 1050--1059.

\bibitem{vovk2005}
V.~Vovk, A.~Gammerman, and G.~Shafer, \emph{Algorithmic Learning in a Random World}.\hskip 1em plus 0.5em minus 0.4em\relax Springer, 2005.

\bibitem{angelopoulos2023conformal}
A.~N. Angelopoulos and S.~Bates, ``Conformal prediction: A gentle introduction,'' \emph{Foundations and Trends in Machine Learning}, vol.~16, no.~4, pp. 494--591, 2023.

\bibitem{stutts2025uncertainty}
A.~C. Stutts, D.~Erricolo, T.~Tulabandhula, M.~Mittal, and A.~R. Trivedi, ``Uncertainty-aware deep reinforcement learning with calibrated quantile regression and evidential learning,'' in \emph{2025 IEEE International Conference on Robotics and Automation (ICRA)}.\hskip 1em plus 0.5em minus 0.4em\relax IEEE, 2025, pp. 9651--9657.

\bibitem{stutts2024mutual}
A.~C. Stutts, D.~Erricolo, S.~Ravi, T.~Tulabandhula, and A.~R. Trivedi, ``Mutual information-calibrated conformal feature fusion for uncertainty-aware multimodal 3d object detection at the edge,'' in \emph{2024 IEEE international conference on robotics and automation (ICRA)}.\hskip 1em plus 0.5em minus 0.4em\relax IEEE, 2024, pp. 2029--2035.

\bibitem{stutts2023lightweight}
A.~C. Stutts, D.~Erricolo, T.~Tulabandhula, and A.~R. Trivedi, ``Lightweight, uncertainty-aware conformalized visual odometry,'' in \emph{2023 IEEE/RSJ International Conference on Intelligent Robots and Systems (IROS)}.\hskip 1em plus 0.5em minus 0.4em\relax IEEE, 2023, pp. 7742--7749.

\bibitem{kumar2025uncertainty}
D.~Kumar, S.~Tayebati, N.~Darabi, V.~P.-H. Hu, and A.~R. Trivedi, ``Uncertainty-aware {LiDAR}-camera autonomy via conformal prediction and principled abstention,'' in \emph{2025 IEEE International Conference on Omni-layer Intelligent Systems (COINS)}.\hskip 1em plus 0.5em minus 0.4em\relax IEEE, 2025, pp. 1--6.

\bibitem{tayebati2025learningconformal}
S.~Tayebati, D.~Kumar, N.~Darabi, D.~Jayasuriya, R.~Krishnan, and A.~R. Trivedi, ``Learning conformal abstention policies for adaptive risk management in large language and vision-language models,'' 2025.

\bibitem{papadopoulos2002inductive}
H.~Papadopoulos, V.~Vovk, and A.~Gammerman, ``Inductive confidence machines for regression,'' \emph{European Conference on Machine Learning}, pp. 345--356, 2002.

\bibitem{lei2018distribution}
J.~Lei, M.~G'Sell, A.~Rinaldo, R.~J. Tibshirani, and L.~Wasserman, ``Distribution-free predictive inference for regression,'' \emph{Journal of the American Statistical Association}, 2018.

\bibitem{darabi2024navigating}
N.~Darabi, P.~Shukla, D.~Jayasuriya, D.~Kumar, A.~C. Stutts, and A.~R. Trivedi, ``Navigating the unknown: Uncertainty-aware compute-in-memory autonomy of edge robotics,'' in \emph{2024 Design, Automation \& Test in Europe Conference \& Exhibition (DATE)}, 2024, pp. 1--6.

\bibitem{trivedi2025intelligent}
A.~R. Trivedi, S.~Tayebati, H.~Kumawat, N.~Darabi, D.~Kumar, A.~K. Kosta, Y.~Venkatesha, D.~Jayasuriya, N.~Jayasinghe, P.~Panda, S.~Mukhopadhyay, and K.~Roy, ``Intelligent sensing-to-action for robust autonomy at the edge: Opportunities and challenges,'' in \emph{2025 Design, Automation \& Test in Europe Conference (DATE)}, 2025, pp. 1--10.

\bibitem{vovk2005algorithmic}
V.~Vovk, A.~Gammerman, and G.~Shafer, \emph{Algorithmic learning in a random world}.\hskip 1em plus 0.5em minus 0.4em\relax Springer, 2005.

\bibitem{romano2019conformalized}
Y.~Romano, E.~Patterson, and E.~Cand{\`e}s, ``Conformalized quantile regression,'' in \emph{Advances in neural information processing systems}, vol.~32, 2019.

\bibitem{martins2016softmax}
A.~Martins and R.~Astudillo, ``From softmax to sparsemax: A sparse model of attention and multi-label classification,'' in \emph{International conference on machine learning}.\hskip 1em plus 0.5em minus 0.4em\relax PMLR, 2016, pp. 1614--1623.

\bibitem{fan2024realtime}
T.~Fan, P.~Chen, X.~Zhao, and J.~Pan, ``Real-time uncertainty-aware motion planning for collaborative robots,'' \emph{Science Robotics}, vol.~9, no.~87, p. eadh1234, 2024.

\bibitem{sun2024probabilistic}
L.~Sun, Z.~Zhou, and C.~Tomlin, ``Probabilistic safety guarantees for learning-based control using conformal prediction,'' in \emph{2024 American Control Conference (ACC)}.\hskip 1em plus 0.5em minus 0.4em\relax IEEE, 2024, pp. 3127--3134.

\bibitem{lee2024highly}
C.~Lee, L.~Rahimifard, J.~Choi \emph{et~al.}, ``Highly parallel and ultra-low-power probabilistic reasoning with programmable gaussian-like memory transistors,'' \emph{Nature Communications}, vol.~15, p. 2439, 2024.

\bibitem{amigoni2018experimental}
F.~Amigoni, N.~Basilico, J.~Banfi, and S.~Ferretti, ``An experimental evaluation of multirobot patrol strategies in realistic environments,'' \emph{Autonomous Robots}, vol.~42, no.~7, pp. 1481--1502, 2018.

\bibitem{karaman2011sampling}
S.~Karaman and E.~Frazzoli, ``Sampling-based algorithms for optimal motion planning,'' \emph{The International Journal of Robotics Research}, vol.~30, no.~7, pp. 846--894, 2011.

\bibitem{glennie2016calibration}
C.~Glennie and D.~D. Lichti, ``Calibration and stability analysis of the velodyne hdl-64e s2 scanner,'' \emph{International Archives of Photogrammetry and Remote Sensing}, vol.~38, no.~3, pp. 219--226, 2016.

\bibitem{geiger2012ready}
A.~Geiger, P.~Lenz, and R.~Urtasun, ``Are we ready for autonomous driving? the kitti vision benchmark suite,'' in \emph{2012 IEEE Conference on Computer Vision and Pattern Recognition}.\hskip 1em plus 0.5em minus 0.4em\relax IEEE, 2012, pp. 3354--3361.

\bibitem{qin2018vins}
T.~Qin, P.~Li, and S.~Shen, ``Vins-mono: A robust and versatile monocular visual-inertial state estimator,'' \emph{IEEE Transactions on Robotics}, vol.~34, no.~4, pp. 1004--1020, 2018.

\bibitem{lin2014microsoft}
T.-Y. Lin, M.~Maire, S.~Belongie, J.~Hays, P.~Perona, D.~Ramanan, P.~Doll{\'a}r, and C.~L. Zitnick, ``Microsoft coco: Common objects in context,'' in \emph{European conference on computer vision}.\hskip 1em plus 0.5em minus 0.4em\relax Springer, 2014, pp. 740--755.

\bibitem{yu2020bdd100k}
F.~Yu, H.~Chen, X.~Wang, W.~Xian, Y.~Chen, F.~Liu, V.~Madhavan, and T.~Darrell, ``Bdd100k: A diverse driving dataset for heterogeneous multitask learning,'' in \emph{Proceedings of the IEEE/CVF Conference on Computer Vision and Pattern Recognition}, 2020, pp. 2636--2645.

\bibitem{cordts2016cityscapes}
M.~Cordts, M.~Omran, S.~Ramos, T.~Rehfeld, M.~Enzweiler, R.~Benenson, U.~Franke, S.~Roth, and B.~Schiele, ``The cityscapes dataset for semantic urban scene understanding,'' in \emph{Proceedings of the IEEE Conference on Computer Vision and Pattern Recognition}, 2016, pp. 3213--3223.

\bibitem{timans2024adaptive}
A.~Timans, C.-N. Straehle, K.~Sakmann, and E.~Nalisnick, ``Adaptive bounding box uncertainties via two-step conformal prediction,'' in \emph{European Conference on Computer Vision}.\hskip 1em plus 0.5em minus 0.4em\relax Springer, 2024, pp. 363--398.

\end{thebibliography}

\end{document}